\documentclass[letterpaper]{article} 
\usepackage{aaai25}   
\usepackage{times}  
\usepackage{helvet}  
\usepackage{courier}  
\usepackage[hyphens]{url}  
\usepackage{graphicx} 
\usepackage{natbib}  
\usepackage{caption} 
\usepackage{algorithm}
\usepackage{algorithmic}
\usepackage{newfloat}
\usepackage{listings}
\DeclareCaptionStyle{ruled}{labelfont=normalfont,labelsep=colon,strut=off} 
\floatstyle{ruled}
\newfloat{listing}{tb}{lst}{}
\floatname{listing}{Listing}
\usepackage{subcaption}
\usepackage{amsmath}
\usepackage{amssymb}
\DeclareMathOperator*{\argmax}{arg\,max}
\DeclareMathOperator*{\argmin}{arg\,min}
\usepackage{cleveref}
\usepackage{booktabs}
\usepackage{multirow}

\usepackage{tabularx}
\usepackage{array}

\title{Enhancing Robustness in Incremental Learning with Adversarial Training}
\author{
    Seungju Cho\equalcontrib,
     Hongsin Lee\equalcontrib,
    Changick Kim\\
}
\affiliations{
    Korea Advanced Institute of Science and Technology (KAIST)\\
    \{joyga, hongsin04, changick\}@kaist.ac.kr
}

\begin{document}

\maketitle

\begin{abstract}
Adversarial training is one of the most effective approaches against adversarial attacks.
However, adversarial training has primarily been studied in scenarios where data for all classes is provided, with limited research conducted in the context of incremental learning where knowledge is introduced sequentially.
In this study, we investigate Adversarially Robust Class Incremental Learning (ARCIL), which deals with adversarial robustness in incremental learning.
We first explore a series of baselines that integrate incremental learning with existing adversarial training methods, finding that they lead to conflicts between acquiring new knowledge and retaining past knowledge.
Furthermore, we discover that training new knowledge causes the disappearance of a key characteristic in robust models: a flat loss landscape in input space.
To address such issues, we propose a novel and robust baseline for ARCIL, named \textbf{FL}atness-preserving \textbf{A}dversarial \textbf{I}ncremental learning for \textbf{R}obustness (\textbf{FLAIR}).
Experimental results demonstrate that FLAIR significantly outperforms other baselines.
To the best of our knowledge, we are the first to comprehensively investigate the baselines, challenges, and solutions for ARCIL, which we believe represents a significant advance toward achieving real-world robustness.
Codes are available at \url{https://github.com/HongsinLee/FLAIR}.

\end{abstract}

\section{Introduction}
\label{sec:intro}

Given the susceptibility of deep neural networks (DNNs) to adversarial attacks, adversarial training is recognized as the most effective defense method \cite{athalye2018obfuscated,wu2020adversarial,wang2021convergence, wu2021wider}.
Furthermore, adversarial training has garnered significant attention not only for its role in improving adversarial robustness but also for enhancing feature representation and interpretation \cite{engstrom2019adversarial,salman2020adversarially,santurkar2020breeds,bai2021clustering,deng2021adversarial,allen2022feature, kireev2022effectiveness}.
Despite its effectiveness, adversarial training has not been sufficiently explored in the context of real-world scenarios with continuously evolving data.
This leads us to address a crucial problem: ensuring robustness in incremental learning environments.
We term this challenge Adversarially Robust Class Incremental Learning (ARCIL), highlighting the need for solutions that combine the strengths of adversarial training with the dynamic requirements of incremental learning.

In incremental learning, a central challenge is acquiring new knowledge without forgetting what has already been learned.
To address this issue, incremental learning employs various methods, such as knowledge distillation \cite{lwf, icarl} and rehearsal techniques \cite{er, der}.
However, these methods do not take adversarial robustness into account, so we construct new baselines for ARCIL by applying adversarial training on those incremental learning in \Cref{tab:cil_adv_frame}.
This includes not only standard adversarial training methods \cite{PGD, TRADES, MART} but also adversarial distillation techniques \cite{ard, rslad, adaad} in place of the traditional knowledge distillation used in incremental learning, to ensure a fair comparison within the framework.
Nonetheless, existing baselines are insufficient to solve ARCIL.

\begin{table*}[ht]
  \centering
 \setlength{\tabcolsep}{2.9pt}
    \fontsize{9pt}{10pt}\selectfont
    \begin{tabularx}{\linewidth}{c|l|l}
    \toprule
    \multicolumn{1}{c|}{\textbf{Type}} & \multicolumn{1}{c|}{\textbf{Methods}} & \multicolumn{1}{c}{\textbf{Formulation}} \\
    \midrule
\multirow{4}{*}{AT} & PGD-AT & $\mathbb{E}_{(\boldsymbol{x}, y)\sim \mathcal{S}_t} \big[l_{CE}(f_t(\boldsymbol{x}_{adv}),y)\big] $\\
\cmidrule{2-3}          & TRADES & $\mathbb{E}_{(\boldsymbol{x}, y)\sim \mathcal{S}_t} \big[l_{CE}(f_t(\boldsymbol{x}),y) + \alpha \cdot l_{KL}(f_t(\boldsymbol{x}_{adv}) \| f_t(\boldsymbol{x}))\big]  $\\
\cmidrule{2-3}          & MART & $\mathbb{E}_{(\boldsymbol{x}, y)\sim \mathcal{S}_t} \big[l_{BCE}(f_t(\boldsymbol{x}_{adv}),\boldsymbol{1}_y) + \alpha \cdot (1- \text{Pr}(f_t(x) = y)) \cdot l_{KL}(f_t(\boldsymbol{x}_{adv}) \| f_t(\boldsymbol{x}))\big]  $\\
 \midrule
\multirow{4}{*}{I-AD} & I-ARD & $\mathbb{E}_{(\boldsymbol{x}, y)\sim \mathcal{S}_t} \big[l_{CE}(f_t(\boldsymbol{x}_{adv}),y) + \beta \cdot l_{KL}([f_t(\boldsymbol{x}_{adv})]^{t-1}_{0} \| f_{t-1}(\boldsymbol{x}))\big] $\\
\cmidrule{2-3}          & I-RSLAD & $\mathbb{E}_{(\boldsymbol{x}, y)\sim \mathcal{S}_t} \big[l_{CE}(f_t(\boldsymbol{x}_{adv}),y) + \beta \!\cdot\! \big(\alpha \!\cdot\! l_{KL}([f_t(\boldsymbol{x}_{adv})]^{t-1}_{0} \| f_{t-1}(\boldsymbol{x})) + (1-\alpha) \!\cdot\! l_{KL}([f_t(\boldsymbol{x})]^{t-1}_{0} \| f_{t-1}(\boldsymbol{x}))\big)\big] $\\
\cmidrule{2-3}          & I-AdaAD & $\mathbb{E}_{(\boldsymbol{x}, y)\sim \mathcal{S}_t} \big[l_{CE}(f_t(\boldsymbol{x}_{adv}),y) + \beta \!\cdot\! \big( \alpha \!\cdot\! l_{KL}([f_t(\boldsymbol{x}_{adv})]^{t-1}_{0} \| f_{t-1}(\boldsymbol{x}_{adv})) + (1-\alpha) \!\cdot\! l_{KL}([f_t(\boldsymbol{x})]^{t-1}_{0} \| f_{t-1}(\boldsymbol{x}))\big)\big]$\\
 \midrule
    \multirow{6}{*}{\shortstack{Non-\\Rehearsal\\R-CIL}} &  R-EWC-on & $\mathbb{E}_{(\boldsymbol{x}, y)\sim \mathit{D_t}}\big[l_{CE}(f_t(\boldsymbol{x}_{adv}),y) + l_{EWC}(\theta_t, \theta_{t-1})\big]$   \\
    \cmidrule{2-3}    & R-LwF & $\mathbb{E}_{(\boldsymbol{x}, y)\sim \mathit{D_t}}\big[l_{CE}(f_t(\boldsymbol{x}_{adv}),y) + \alpha \cdot l_{KL}([f_t(\boldsymbol{x})]^{t-1}_{0} \| f_{t-1}(\boldsymbol{x}))\big]  $ \\
\cmidrule{2-3}          & R-LwF-MC & $\mathbb{E}_{(\boldsymbol{x}, y)\sim \mathit{D_t}}\big[l_{BCE} ([f_t(\boldsymbol{x}_{adv})]^{t}_{t-1},\boldsymbol{1}_y) \ +  l_{BCE}([f_{t}(\boldsymbol{x})]^{t-1}_{0}, {f}_{t-1}(\boldsymbol{x}))\big]$   \\
\cmidrule{2-3}& R-SI & $\mathbb{E}_{(\boldsymbol{x}, y)\sim \mathit{D_t}}\big[l_{CE}(f_t(\boldsymbol{x}_{adv}),y) + l_{SI}(\theta_t, \theta_{t-1})\big] $   \\
\midrule  
\multirow{9}{*}{\shortstack{Rehearsal\\R-CIL}}
& R-ER & $  \mathbb{E}_{(\boldsymbol{x}, y)\sim \mathit{D_t}}\big[l_{CE}(f_t(\boldsymbol{x}_{adv}),y)\big] + \mathbb{E}_{(\boldsymbol{x}, y)\sim \mathit{M}} \big[l_{CE}(f_t(\boldsymbol{x}_{adv}),y)\big]  $   \\
\cmidrule{2-3}          & R-ER-ACE & $  \mathbb{E}_{(\boldsymbol{x}, y)\sim \mathit{D_t}}\big[l_{ACE}(f_t(\boldsymbol{x}_{adv}),C_{curr})\big] + \mathbb{E}_{(\boldsymbol{x}, y)\sim \mathit{M}} \big[l_{CE}(f_t(\boldsymbol{x}_{adv}),y)\big] $   \\
\cmidrule{2-3}          & R-DER & $  \mathbb{E}_{(\boldsymbol{x}, y)\sim \mathit{D_t}}\big[l_{CE}(f_t(\boldsymbol{x}_{adv}),y)\big] + \mathbb{E}_{(\boldsymbol{x}, \mathbf{z})\sim \mathit{M}} \big[ \alpha \cdot 
 l_{MSE}(f_t(\boldsymbol{x}_{adv}),\mathbf{z})\big] $ \\
\cmidrule{2-3}          & R-DER++ & $  \mathbb{E}_{(\boldsymbol{x}, y)\sim \mathit{D_t}}\big[l_{CE}(f_t(\boldsymbol{x}_{adv}),y)\big] + \mathbb{E}_{(\boldsymbol{x}, \mathbf{z}, y) \sim \mathit{M}} \big[\alpha \cdot l_{MSE}(f_t(\boldsymbol{x}_{adv}),\mathbf{z}) + \beta \cdot l_{CE}(f_t(\boldsymbol{x}_{adv}),y)\big] $\\
\cmidrule{2-3}          & R-iCaRL & $\mathbb{E}_{(\boldsymbol{x}, y)\sim \mathit{D_t} \cup B_{t-1}}\big[l_{BCE} ([f_t(\boldsymbol{x}_{adv})]^{t}_{t-1},\boldsymbol{1}_y) \ +  l_{BCE}([f_{t}(\boldsymbol{x})]^{t-1}_{0}, {f}_{t-1}(\boldsymbol{x}))\big]$ \\
\cmidrule{2-3}          & R-LUCIR   &  $\mathbb{E}_{(\boldsymbol{x}, y)\sim \mathit{D_t} \cup B_{t-1}}\big[l_{CE} (f_t(\boldsymbol{x}_{adv}),y) + \alpha \cdot l_{dis}^G(\boldsymbol{x}_{adv}) \big] \ + \mathbb{E}_{(\boldsymbol{x}, y)\sim B_{t-1}}\big[ \beta \cdot l_{mr}({\boldsymbol{x}_{adv}})\big]$ \\

\midrule
\multirow{2.5}{*}{ARCIL} & TABA & $\mathbb{E}_{(\boldsymbol{x}, y)\sim \mathit{D_t} \cup B_{t-1} \cup A_\text{TABA}}\big[l_{BCE} ([f_t(\boldsymbol{x}_{adv})]^{t}_{t-1},\boldsymbol{1}_y) \ +  l_{BCE}([f_{t}(\boldsymbol{x}_{adv})]^{t-1}_{0}, {f}_{t-1}(\boldsymbol{x}))\big]$ \\
\cmidrule{2-3}          & \textbf{FLAIR} & $\mathbb{E}_{(\boldsymbol{x}, y)\sim S_t}\big[l_{BCE} ([f_t(\boldsymbol{x}_{adv})]^{t}_{t-1},\boldsymbol{1}_y) + \alpha \cdot l_{BCE}([f_{t}(\boldsymbol{x}_{adv})]^{t-1}_{0}, {f}_{t-1}(\boldsymbol{x}_{adv}))  
+ \beta \cdot l_{FPD}(\boldsymbol{x}, \boldsymbol{x}_{adv} ; f_t, f_{t-1})\big]$\\
    \bottomrule
    \end{tabularx}
    \caption{Different methods to fit in ARCIL setting. Type AT stands for naive Adversarial Training on ARCIL, while I-AD is revised adversarial distillation methods by considering the previous task model as a teacher model.
R-CIL consists of a set of revised CIL methods, mainly by changing the input $\boldsymbol{x}$ of the learning incremented task to $\boldsymbol{x}_{adv}$. R-CIL is further categorized into two subtypes: Non-Rehearsal and Rehearsal types.
Details of specific notations can be found in the appendix.}
  \label{tab:cil_adv_frame}%
\end{table*}

We identify several reasons why the baseline methods struggle to perform well.
Firstly, the straightforward application of adversarial distillation leads to conflicts between learning new tasks and preserving knowledge from previous tasks.
This issue is exacerbated in adversarial training, due to its adversarial input.
Secondly, we figure out that incorporating new tasks causes the disappearance of one of the main characteristics of adversarial training, a flat loss landscape \cite{qin2019adversarial, curvature}.
Given the abundance of research on adversarial robustness and flatness, we believe preserving flatness is important to maintain robustness.
However, we observe in \Cref{tab:cil_adv_frame} that the baselines tend to forget the flatness of the loss for past tasks as new tasks are introduced.
Lastly, ARCIL inherently suffers from a lack of training data, which is a significant issue since adversarial training generally requires more data than natural training \cite{rebuffi2021fixing, li2023data, yue2024revisitingLT}.

To resolve these issues, we propose a  \textbf{FL}atness-preserving \textbf{A}dversarial \textbf{I}ncremental learning for \textbf{R}obustness (\textbf{FLAIR}).
The proposed method employs separated-logits \cite{icarl, ahn2021ss} to ensure that the logits corresponding to past tasks remain unaffected when learning new tasks.
This approach mitigates conflicts between retaining existing knowledge and acquiring new information while leveraging the benefits of adversarial distillation.
Additionally, FLAIR provides a straightforward method for preserving loss flatness.
By employing Taylor approximation, we observe that the output difference between clean and adversarial example, encapsulates gradient and hessian information.
Preserving this output difference ensures that gradient and hessian details are retained, thereby maintaining the flatness.
Furthermore, we employ data augmentation techniques to address the issue of insufficient training data.

In summary, our contributions are as follows:

\begin{itemize}
\item We address the less well-studied problem of ARCIL and propose baselines that incorporate both incremental learning and adversarial training.
\item 
We systematically analyze three key issues with existing baselines for ARCIL and propose FLAIR, which provides effective solutions to address these challenges.
\item We evaluate the performance in the ARCIL setting, and FLAIR surpasses baseline results, achieving superior performance.
\end{itemize}

\section{Related Work}
In this section, we review adversarial training and incremental learning approaches in the context of image classification.

\subsection{Adversarial Training}

Adversarial training improves model robustness by addressing the following min-max problem.
\begin{equation}\label{eq:basic_adv}
\begin{split}
    \argmin_{\theta}  &\mathbb{E}_{(\boldsymbol{x}, y)\sim \mathit{D}}\left [    l_{min}\left (f_{\theta}(\boldsymbol{x}_{adv}), y \right ) \right  ], \ \\
    \text{where} &\ \boldsymbol{x}_{adv} = \argmax_{ \| \boldsymbol{x}_{adv} - \boldsymbol{x} \|_p \leq \epsilon}\ l_{max}\left (f_{\theta}(\boldsymbol{x}_{adv}), y \right ),
\end{split}
\end{equation}
where $\mathit{D}$ is the data distribution, $\theta$ is parameters of the model $f$, $\epsilon$ is the maximum perturbation limit, and $l_{min}$ and $l_{max}$ denote loss functions for min-max problem.
Most adversarial training uses a multi-step based Projected Gradient Attack (PGD) \cite{PGD} for the inner maximization problem, and several regularization losses have been proposed to solve the outer minimization problem of \eqref{eq:basic_adv}.
Representatively, TRADES \cite{TRADES} incorporates the KL divergence loss between the output of clean and adversarial images, and MART \cite{MART} introduces per-sample weights through the confidence of each sample.

To enhance the effectiveness of adversarial training, various techniques have been proposed, including data augmentation \cite{rebuffi2021fixing, li2023data} and knowledge distillation \cite{ard, rslad, iad, akd,zhao2022enhanced, adaad}.
Among them, methods such as ARD \cite{ard}, RSLAD \cite{rslad}, and AdaAD \cite{adaad} use knowledge distillation to transfer the robustness of large teacher models.
These adversarial distillation approaches significantly improve the performance of small models when a robust teacher model is available.

A flat loss landscape is closely related to generalization, and therefore, it impacts adversarial robustness performance \cite{izmailov2018averaging, chen2020robust}.
Although several studies have proposed methods to achieve a flatter loss landscape \cite{curvature,qin2019adversarial}, most research on adversarial training has focused on scenarios where all data is provided upfront. 
In incremental settings, where data is received sequentially and past data may be lost, maintaining a flatter loss function becomes challenging as new data is introduced and old data is not available. 
Our research investigates methodologies for achieving adversarial robustness in such incremental settings while addressing the preservation of a flatter loss function.

\subsection{Class Incremental Learning}

Class incremental learning (CIL) assumes data from new, non-overlapping classes is provided sequentially.
The distribution shift leads to a phenomenon known as \textit{catastrophic forgetting} \cite{CatastrophicForgetting}, where the model forgets previously learned information as it learns new data.
The central challenge in CIL is to integrate new data effectively while preserving knowledge from earlier tasks.

Various methods have been proposed to tackle this challenge, including knowledge distillation \cite{lwf, icarl, ahn2021ss}, weight regularization \cite{ewc, lwf, synaptic, online-ewc}, or memory buffer \cite{icarl, er, lucir, der, er_ace}.
Among these, memory buffer-based approaches, particularly rehearsal methods, have become strong baselines due to their effectiveness and compatibility with other techniques \cite{icarl,der,er,lucir,er_ace}.
In contrast, non-rehearsal methods have been extensively studied because they are more applicable to real-world scenarios \cite{boschini2022class, aljundi2019gradient, farquhar2018towards}.


While many CIL methods aim to mitigate catastrophic forgetting, there is a notable research gap concerning adversarially perturbed data. Recent studies have explored adversarial robustness in incremental learning but have primarily relied on simple data augmentation techniques and may not have fully addressed the complexities of the problem \cite{bai2023towards}. 
Our work addresses this gap by offering a more robust and comprehensive examination of the impact of adversarial training.

\section{Proposed Method}
We first analyze the underlying issues of ARCIL and then introduce our proposed method.

\subsection{Problem Formulation}
A model $f$ is sequentially trained on tasks $1, 2, \dots, T$, with $f_t$ denoting the model trained on task $t$.
In each individual task $t$, inputs $\boldsymbol{x} \in \mathbb{R}^{d} $ and labels $y \in \mathbb{R}$ are independently and identically drawn from the task-specific data distribution $D_t = \{X_t, Y_t\}$ , where $D_i \cap D_j = \emptyset$ for $i\neq j$.
Here, $X_t$ and $Y_t$ are the set of inputs and true labels of task $t$.
The goal is to maintain robustness against adversarial attacks in new tasks while preserving the robustness learned in previous tasks.

We investigate two incremental learning settings.
The first setting involves the absence of any data from previous tasks (Non-Rehearsal), while the second setting deals with having a small amount of data from previous tasks (Rehearsal).
For the setting with a rehearsal buffer, we utilized a limited and constant memory size for storing past data, as proposed in previous works such as \cite{icarl, er, lucir,der, bai2023towards}.
In our method, the rehearsal buffer $B_{t-1}$ is selected with the herding algorithm after training on each task as in \cite{icarl}, and concatenate the buffer at the beginning of each incremental task on the current dataset, denoted as $D_t \cup B_{t-1}$.
Regardless of using the rehearsal buffer, we denote the current training dataset as $S_t$, instead of $D_t$ or $D_t \cup B_{t-1}$ if there is no misunderstanding.

\subsection{Adversarial Distillation with Separated Logit}
Knowledge distillation is an effective technique for achieving high performance in adversarial training by imparting robustness from a strong teacher model \cite{ard,adaad}.
Similarly, knowledge distillation can be utilized in incremental learning by leveraging a model trained on past tasks as a teacher \cite{lwf,icarl}.
Thus, intuitively incorporating these two methods to address ARCIL can provide a reasonable solution.
\begin{equation}
\label{eq:method-1}
 \begin{aligned}
   l(\boldsymbol{x},y) = l_{CE}(f_t(\boldsymbol{x}_{adv}),y)
    + \alpha \cdot l_{AD}(\boldsymbol{x},\boldsymbol{x}_{adv};f_t, f_{t-1})),
\end{aligned}   
\end{equation}
where $\alpha$ is a hyperparameter that controls the strength of knowledge distillation from the past task model to the current model, and $l_{AD}$  represents the adversarial distillation loss function, such as ARD \cite{ard}, RSLAD \cite{rslad}, or AdaAD \cite{adaad}.
However, naively applying adversarial distillation to incremental learning decreases overall performance as in \Cref{tab:main_res18_nobuffer} and \Cref{tab:main_res18_buffer}.

We attribute the insufficient performance to the conflicting
objectives of learning new knowledge and maintaining
previous robustness.
This represents a common trade-off in incremental learning known as the stability-plasticity dilemma \cite{mermillod2013stability}, but is more exacerbated in ARCIL because of adversarial input, $\boldsymbol{x}_{adv}$.
In \Cref{eq:basic_adv}, the adversarial input maximizes the inner loss function, leading to new task data being adversarially crafted, often increasing the logit of one of the previous task classes.
In such cases, reducing the attacked logit of the adversarial input is necessary to learn the new task, but this can result in even more severe forgetting than in natural incremental learning.
In contrast, distillation aims to preserve the logit of previous tasks.
This conflict between the two loss functions explains why naively applying adversarial distillation does not work well in ARCIL.

\begin{table*}[t]
 \centering
 \setlength{\tabcolsep}{4.6pt}
    \fontsize{9pt}{10pt}\selectfont
 \begin{tabular}{clrrrrrrrrrrrr}

 \toprule
 \multicolumn{1}{c}{\multirow{2.5}{*}{Type}} &\multicolumn{1}{c}{\multirow{2.5}{*}{Method}} & \multicolumn{4}{c}{S-CIFAR10}& \multicolumn{4}{c}{S-CIFAR100}& \multicolumn{4}{c}{S-SVHN}\\
  \cmidrule(l){3-6}
\cmidrule(l){7-10}
 \cmidrule(l){11-14}& & Clean$\uparrow$ & PGD$\uparrow$ & AA$\uparrow$ & R-BWT$\uparrow$ & Clean$\uparrow$ & PGD$\uparrow$ & AA$\uparrow$ & R-BWT$\uparrow$ & Clean$\uparrow$ & PGD$\uparrow$ & AA$\uparrow$ & R-BWT$\uparrow$  \\
\midrule
\multirow{3}{*}{AT} &PGD-AT & 18.98& 16.30& 16.27&  -72.36& 8.14& 4.82& 4.73&  -39.54  & 19.10 & 10.72 & 6.45 & -73.80 \\
  &TRADES & 18.26& 15.83& 15.81&  -70.45 & 8.27& 5.78& 5.61&  -48.32 &
 19.07 & 11.02 & 6.78 &-76.96\\
  &MART & 18.78& 16.37& 16.32&  -73.53 & 8.19& 5.60& 5.35& -47.36 & 18.47 & 11.50 & 6.90 & -76.22\\

\midrule

\multirow{3}{*}{I-AD}  &I-ARD& 19.03& 16.33& 16.32&  -72.81& 8.28& 5.28& 5.14&   -44.06 & 20.81 & 
10.07 & 9.10 & -71.73 \\
  &I-RSLAD& 18.84& 16.10& 16.10&  -71.80& 8.38& 5.29& 5.15&   -44.50 & 19.07  & 10.42 & 6.31 & -71.95 \\
  &I-AdaAD& 18.89& 16.19& 16.17&  -71.90& 8.46& 5.39& 5.12&   -43.80 & 18.99 & 10.70 & 6.49 & -72.15 \\
\midrule

 \multirow{4}{*}{R-CIL}  &R-LwF & 18.98& 16.31& 16.28&  -72.66& 8.28& 5.08& 5.03&   -42.56 & 19.16 & 10.31 & 6.22 & -72.32 \\
  &R-LwF-MC & 40.24& 15.85& 14.77&  -34.05 & 27.16& 8.16& 6.54& -19.68 & 58.02 & 5.01 & 3.20 & -42.98 \\
  &R-EWC-on & 18.08& 8.47& 8.16&  -58.98& 6.62 & 2.78& 2.68&   -19.74 & 11.12 & 7.02 & 4.65 & -63.86 \\
  &R-SI & 18.40& 15.54& 15.50&  -67.78& 8.45 & 5.40& 5.26&   -45.52 & 18.50 & 11.15 & 6.65 & -73.53\\

\midrule

\multirow{2}{*}{ARCIL}  &\textbf{FLAIR}& \underline{43.64}& \underline{18.42}& \underline{17.02}&  \textbf{-22.09}&  \underline{28.72}& \underline{11.02}& \underline{8.59}&  \underline{-15.06} & \underline{65.16}& \underline{16.25} & \underline{15.96} & \textbf{-16.86} \\
   &\textbf{FLAIR+}& \textbf{44.22}& \textbf{22.42}& \textbf{20.46}& \underline{-24.66}& \textbf{29.47}& \textbf{13.60}& \textbf{10.20}& \textbf{-14.26} & \textbf{66.26} & \textbf{30.04} & \textbf{27.21} & \underline{-17.53}\\
 \bottomrule
 \end{tabular}

   \caption{Clean, 20-step PGD, AutoAttack (AA) accuracy (\%), and Robust Backward Transfer (R-BWT) measured on ResNet-18 for S-CIFAR10, S-CIFAR100, and S-SVHN without a memory buffer.}
    \label{tab:main_res18_nobuffer}
\end{table*}

To prevent this conflict, we design the loss function to avoid affecting the logits of past classes when learning new tasks, using binary cross-entropy loss only on the new class indices.
This simple technique not only prevents conflicts but also allows for greater flexibility in the distillation process.
In summary, in the incremental setting, we apply adversarial distillation with substitution of cross entropy $l_{CE}(\boldsymbol{x}_{adv},y)$ into binary cross-entropy loss with current task logit $l_{BCE} ([f_t(\boldsymbol{x}_{adv})]^{t}_{t-1},\boldsymbol{1}_y)$.
Here, $[\cdot]_{t-1}^t$ represents a slicing operation that returns the outputs immediately after the $(t-1)$-th task up to and including the $t$-th task output, and $\boldsymbol{1}_y$ is one-hot vector for label $y$.
To preserve past knowledge through distillation, methods such as ARD, RSLAD, or AdaAD can be used for $l_{AD}(\boldsymbol{x},\boldsymbol{x}_{adv};f_t,f_{t-1})$.
We experimentally select AdaAD for our method as follows: $ l_{BCE}([f_{t}(\boldsymbol{x}_{adv})]^{t-1}_{0}, {f}_{t-1}(\boldsymbol{x}_{adv}))$.

Overall, we define the adversarial distillation with separated logit (ADSL) as follows:
\begin{equation}\label{ADSL_loss}
\begin{split}
   l_{ADSL}(\boldsymbol{x},y) &= l_{BCE} ([f_t(\boldsymbol{x}_{adv})]^{t}_{t-1},y) \\
   &+ \alpha \cdot l_{BCE}([f_{t}(\boldsymbol{x}_{adv})]^{t-1}_{0}, {f}_{t-1}(\boldsymbol{x}_{adv})),
\end{split}
 \end{equation}
 where $\alpha$ is a hyperparameter that controls the weight of preserving the previous model's knowledge.

\subsection{Flatness Preserving Distillation} 
To further improve the capability of retaining knowledge, we investigate whether adversarially robust models maintain their distinctive characteristics while learning new tasks.
We focus on the flatness of the loss landscape, a characteristic of robust models \cite{curvature, qin2019adversarial}, and assess it using gradient and Hessian calculations.
We define Gradient Forgetting (GF) and Hessian Forgetting (HF) to evaluate the preservation of loss flatness.
\begin{equation}
\begin{aligned}
    \text{GF} &= \frac{1}{T-1} \sum_{i = 1}^{T-1} 
\mathbb{E}_{(\boldsymbol{x},y)\sim D_i} \big[ \left\| \nabla f_T(\boldsymbol{x}) - \nabla f_{i}(\boldsymbol{x}) \right \|_2 \big],\\
    \text{HF} &= \frac{1}{T-1} \sum_{i = 1}^{T-1} \mathbb{E}_{(\boldsymbol{x},y)\sim D_i} \big[ \left\| \mathbf{H}_T(\boldsymbol{x}) - \mathbf{H}_i(\boldsymbol{x})) \right \|_F\big], 
\end{aligned}
\label{eqn:GFHF}
\end{equation}
where $D_i$ is $i$-th task test dataset, and $\mathbf{H}_i$ denotes hessian matrix of $i$-th task learned model.
We find that all baselines in \Cref{tab:cil_adv_frame} exhibit high GF and HF, indicating a disappearance of flatness, as shown in \Cref{tab:GFHF}.
This suggests that the loss landscape learned through adversarial training for each task is altered, which we hypothesize is linked to robustness forgetting.
Since the flatness of the loss landscape disappears after learning new tasks, we term this phenomenon the \textit{flatness forgetting} problem.

To address this problem, we consider the gradient and hessian indirectly, as direct calculation of the gradient and hessian is computationally expensive. 
For a clean input $\boldsymbol{x}$ from the data of the current training dataset $S_t$, the adversarial perturbed input is given by:
\begin{equation}\label{method-1}
\boldsymbol{x}_{adv} =   \boldsymbol{x} +    \boldsymbol{\delta}.
\end{equation}
For a small perturbation $\boldsymbol{\delta}$, the output of $\boldsymbol{x}_{adv}$ given the $t$-th task model $f_t$ can be approximated using Taylor expansion through the input space.

\begin{equation}
\begin{split}
    f_{t}(\boldsymbol{x}_{adv}) = f_{t}(\boldsymbol{x}) + 
    \nabla f_t(\boldsymbol{x})^{T} \boldsymbol{\delta} 
    + \frac{1}{2}\boldsymbol{\delta}^T \mathbf{H}_{t}(\boldsymbol{x}) \boldsymbol{\delta},
\end{split}
\label{method-2}
\end{equation}
where $\nabla f_t(\boldsymbol{x})$ and $\mathbf{H}_t$ are the gradient and hessian matrix of the $t$-th task model, and we neglect higher order terms.
Similarly, the output of the $(t-1)$-th task model can also be approximated using Taylor expansion given the $t$-th task input $\boldsymbol{x}_{adv}$.

Maintaining the first-order and second-order terms in \eqref{method-2} between $t$ and $(t-1)$-th task model is necessary to preserve the flatness. 
Therefore, we exploit the subtraction between clean and adversarial outputs as follows:

\begin{equation}\label{method-5}
\begin{split}
    \Delta f_t 
    &= \nabla f_{t}(\boldsymbol{x})^{T} \boldsymbol{\delta} 
    + \frac{1}{2} \boldsymbol{\delta}^T \mathbf{H}_{t}(\boldsymbol{x}) \boldsymbol{\delta},
\end{split}
\end{equation}
where $\Delta f_t  = f_t( \boldsymbol{x}_{adv}) - f_t(\boldsymbol{x}) $.
Here, we neglect the high-order term due to the small magnitude of the perturbation $\boldsymbol{\delta}$.
As a result, we can efficiently distill both the gradient and hessian information of the $(t-1)$-th task model into $t$-th task model by using the difference between clean and adversarial inputs:
\begin{equation}\label{FPD_loss}
\begin{split}
    l_{FPD}(\boldsymbol{x}, \boldsymbol{x}_{adv} ; f_t, f_{t-1}) = D\left (\Delta f_t \ , \Delta f_{t-1} \right ),
\end{split}
 \end{equation}
where $D$ is a difference metric such as the KL divergence loss.

\subsection{Augmentation Training Data}
Lastly, we highlight another challenge for ARCIL: the lack of training data in each task.
It is well known that AT requires a large amount of training data to capture richer representations for robustness sufficiently \cite{rebuffi2021fixing, li2023data, yue2024revisitingLT}.
However, in ARCIL, the number of data points from previous and future tasks is minimal or nonexistent, unlike in standard AT.
A widely used and effective technique for enhancing features in AT is data augmentation. 
Although TABA \cite{bai2023towards} adopts Mixup augmentation on the selected buffer samples, we find that simple yet effective augmentation is sufficient to enhance robustness in ARCIL.

\citeauthor{yue2024revisitingLT} asserts that utilizing a broader range of augmentations beyond the commonly used horizontal flip would be more effective in long-tailed distributions, which often lack sufficient training data for specific classes.
The rationale behind this is that while traditional AT benefits from a wealth of data for each class, making simple augmentations sufficient, the lack of training data in the long-tailed distributions is insufficient for robust learning.
The rationale behind this is that while traditional AT benefits from a wealth of data for each class, making simple augmentations sufficient, the lack of training data in the long-tailed distributions is insufficient for robust learning.
Similarly, ARCIL scenarios have partial access to past data, thus requiring more diverse augmentations would be effective.
To enhance performance, we follow the methods proposed by \citeauthor{yue2024revisitingLT} by introducing RandAugment (RA) \cite{cubuk2020randaugment} or AutoAugment (AuA) \cite{cubuk2019autoaugment}.

\subsection{Flatness-preserving Adversarial Incremental learning for Robustness}

We propose a novel method for ARCIL, naming our technique \textit{\textbf{FL}atness-preserving \textbf{A}dversarial \textbf{I}ncremental \textbf{L}earning for \textbf{R}obustness} (\textbf{FLAIR}). 
 Our training loss function is as follows.
 \begin{align*}
   l_{FLAIR}(\boldsymbol{x},y) &= l_{BCE} ([f_t(\boldsymbol{x}_{adv})]^{t}_{t-1},y) \\
   &+ \alpha \cdot l_{BCE}([f_{t}(\boldsymbol{x}_{adv})]^{t-1}_{0}, {f}_{t-1}(\boldsymbol{x}_{adv})) \\
&+ \beta \cdot l_{FPD}(\boldsymbol{x}, \boldsymbol{x}_{adv} ; f_t, f_{t-1})
\end{align*}
where $\alpha$ and $\beta$ are hyper-parameter.
Additionally, when we train our method with augmentations such as RA or AuA, we denote it as \textbf{FLAIR+}.

\begin{table*}[t]
 \centering
 \setlength{\tabcolsep}{4.7pt}
    \fontsize{9pt}{10pt}\selectfont
 \begin{tabularx}{\linewidth}{clrrrrrrrrrrrr}

 \toprule

   \multicolumn{1}{c}{\multirow{2.5}{*}{Type}} &\multicolumn{1}{c}{\multirow{2.5}{*}{Method}} & \multicolumn{4}{c}{S-CIFAR10}& \multicolumn{4}{c}{S-CIFAR100} & \multicolumn{4}{c}{S-SVHN}\\
 \cmidrule(l){3-6}
\cmidrule(l){7-10}
 \cmidrule(l){11-14} &
   & Clean$\uparrow$ & PGD$\uparrow$ & AA$\uparrow$ & R-BWT$\uparrow$& Clean$\uparrow$ & PGD$\uparrow$ & AA$\uparrow$ & R-BWT$\uparrow$  & Clean$\uparrow$ & PGD$\uparrow$ & AA$\uparrow$ &R-BWT$\uparrow$  \\
\midrule

\multirow{3}{*}{AT}  &PGD-AT & 52.73& 24.34& 23.86&  -59.35& 19.32& 7.33& 7.20&  -29.24  & 62.76& 22.54& 18.62&-56.38\\
   &TRADES & 40.29& 22.99& 22.03&  -63.16& 17.76& 9.03& 8.68&   -38.90 & 49.83& 20.53& 16.70&-63.91\\
   &MART & 51.08& 28.26& 26.09&  -59.40& 19.67& 8.44& 8.18&   -34.99 & 60.47& 23.79& 19.89&-52.93\\
\midrule

 \multirow{3}{*}{I-AD}  &I-ARD & 50.52& 26.99& 26.12&  -57.00& 18.84& 8.14& 7.94&   -36.69 & 56.06& 20.53& 15.79&-57.95\\
   &I-RSLAD& 49.89& 27.29& 26.17&  -56.60& 18.72& 7.96& 7.64&   -36.02 & 53.39& 19.17& 14.19&-58.85\\
   &I-AdaAD& 52.06& 27.98& 27.10&  -56.16& 20.19& 8.07& 7.66&   -35.03 & 56.25& 20.89& 15.78&-57.13\\
\midrule

 \multirow{6}{*}{R-CIL}  &R-ER & 52.03& 24.23& 23.85& -58.04& 18.09& 7.00& 6.82&  -30.21& 57.76& 21.00& 17.51&-57.92\\

  &R-ER-ACE & \underline{62.93}& 20.17& 19.57& -23.05& 31.87& 7.00& 6.74&  -15.14 & 78.36& 27.26& 26.68&-25.29\\
   &R-DER & 24.30& 16.21& 16.10& -61.24& 16.13& 7.13& 6.06&  -38.82 & 33.17& 11.66& 8.08&-66.56\\ 
   &R-DER++ & 27.75& 16.27& 16.02& -70.53& 19.78& 7.39& 6.95&  -31.43 & 32.27& 11.29& 7.50&-67.11\\ 
   &R-iCaRL & 57.78& 27.29& 25.87& -13.51& 36.47& 9.73& 8.20&  -17.92 & 76.47& 18.02& 17.13&-38.59\\ 
   &R-LUCIR & 62.90& 29.23& 26.12& -30.71& 23.88& 8.80& 7.74&  -24.91 & 61.04& 25.38& 20.28&-51.45\\
\midrule
 \multirow{3}{*}{ARCIL}   &TABA  & 59.94& 25.31& 24.18&  -18.75& 28.44& 8.02& 7.16&   -22.06 & 60.30& 16.81& 12.03&-35.12\\

   &\textbf{FLAIR}&  \textbf{63.81}& \underline{30.28}& \underline{27.65}&  \textbf{-12.08}& \textbf{38.66}& \underline{17.04}& \underline{13.45}& \textbf{-7.87} & \underline{80.83}& \underline{31.11}& \underline{29.77}&\textbf{-1.58}\\
  &\textbf{FLAIR+}& 61.29& \textbf{33.35}& \textbf{30.06}& \underline{-12.49}& \underline{38.04}& \textbf{18.03}& \textbf{14.43}&  \underline{-8.43} & \textbf{83.83}& \textbf{44.33}& \textbf{40.10}& \underline{-2.74}\\
 \bottomrule
 \end{tabularx}
     \caption{Clean, 20-step PGD, AutoAttack (AA) accuracy (\%), and Robust Backward Transfer (R-BWT) measured on ResNet-18 for S-CIFAR10, S-CIFAR100, and S-SVHN with 2000 size of memory buffer.}
 \label{tab:main_res18_buffer}
\end{table*}

\section{Experiments}

\subsection{Experimental Settings}
\subsubsection{Datasets} 
We conducted experiments with and without a memory buffer on the following datasets.
\textit{Split CIFAR-10 (S-CIFAR10)} divides  CIFAR-10 \cite{krizhevsky2009learning} into five tasks, each consisting of two classes.
\textit{Split CIFAR-100 (S-CIFAR100)} divides CIFAR-100 \cite{krizhevsky2009learning} into ten tasks, with ten classes per task.
\textit{Split SVHN (S-SVHN)} divides the SVHN \cite{SVHN} into five tasks, with two classes per task.
Additionally, \textit{Split TinyImageNet (S-TinyImageNet)}, which divides Tiny ImageNet \cite{le2015tiny} into ten tasks with 20 classes per task using a memory buffer, can be found in the appendix.

\subsubsection{Training}
For adversarial training, we used ten steps of PGD attack with a random start, and the maximum perturbation is limited to $\epsilon_{\infty} = 8/255$, while each step is taken with a step size of $2/255$.
In all experiments, we utilized the ResNet-18 architecture \cite{he2016deep}, with additional results on MobileNetV2 \cite{sandler2018mobilenetv2} provided in the appendix.
We conducted a grid search over the hyperparameters \{0,0.5,1,2,4\} for our method and all baseline methods, reporting the best results.

\subsubsection{Baseline}
We devise four type baselines for solving ARCIL to discuss our proposed methods fairly: AT, AD, R-CIL, and ARCIL in \Cref{tab:cil_adv_frame}.
AT represents naive adversarial training methods including PGD-AT \cite{PGD}, TRADES \cite{TRADES}, and MART \cite{MART} on ARCIL without considering any techniques for preventing forgetting.
I-AD is revised adversarial distillation methods with learning new tasks for ARCIL by considering the previous task model as a teacher model, including I-ARD \cite{ard}, I-RSLAD \cite{rslad}, and I-AdaAD \cite{adaad}, where the prefix "I-" indicates modifications for the incremental setting derived from the original methods.
R-CIL consists of adversarially trained CIL baselines using both Non-Rehearsal and Rehearsal methods. For Non-Rehearsal R-CIL, the methods include R-EWC-on \cite{ewc}, R-LwF \cite{lwf}, R-LwF-MC \cite{icarl}, and R-SI \cite{synaptic}, while for Rehearsal R-CIL, the methods include R-ER \cite{er}, R-ER-ACE \cite{er_ace}, R-DER/DER++ \cite{der}, R-iCaRL \cite{icarl}, and R-LUCIR \cite{lucir}, where the prefix "R-" indicates modifications to enhance robustness in CIL methods.
We implement TABA\footnotemark for type ARCIL.
Detailed settings are in the appendix.

\footnotetext{No open-source code is available for TABA. As a result, our results may differ from those reported for the original TABA. For further comparison with TABA results, please refer to the appendix.}

\noindent \textbf{Evaluation}
We measured clean, 20 steps of PGD, and AutoAttack (AA) \cite{AutoAttack} accuracy after learning all incremental tasks with $\epsilon_{\infty} = 8/255$.
To measure forgetting on robustness, we consider the robust backward transfer (R-BWT) metric as follows. 
\begin{equation}
    \text{R-BWT} = \frac{1}{T-1}\sum_{t = 1}^{T-1} 
({RA}_{T,t} - {RA}_{t,t}),
\end{equation}
where ${RA}_{i,j}$ denotes PGD accuracy of task $j$ after learning task $i$.
Accuracy of past task usually drops as new tasks are learned, so a higher R-BWT means less forgetfulness.

\subsection{Main Results}
In \Cref{tab:main_res18_nobuffer} and \Cref{tab:main_res18_buffer}, we present a summary of the experimental results for all methods.
The low performance of type AT indicates that standard AT is not appropriate for ARCIL.
The insufficient performance of type I-AD indicates that distillation methods without considering separated logits result in high R-BWT, failing to resolve the issue of forgetting.
Moreover, most methods that perform well in standard CIL show significantly lower performance, indicating that simply applying AT to CIL is insufficient for addressing ARCIL.
Our methods achieve the highest clean and robust accuracy across all datasets, demonstrating the effectiveness of our approach.
Specifically, R-BWT, which measures the degree of forgetting, also achieved the highest score, indicating that our method effectively mitigates forgetting.
\begin{table}[hb]
 \centering
\setlength{\tabcolsep}{9.6pt}
\fontsize{9pt}{10pt}\selectfont
 \begin{tabularx}{\columnwidth}{lrrrr}

 \toprule

 \multicolumn{1}{c}{\multirow{2}{*}{Method}} & \multicolumn{2}{c}{S-CIFAR10}& \multicolumn{2}{c}{S-CIFAR100}\\
 \cmidrule(l){2-3}
\cmidrule(l){4-5} & GF$\downarrow$ & HF$\downarrow$ & GF$\downarrow$ & HF$\downarrow$ \\
\midrule
 PGD-AT & 1.641& 2.184 &  1.832& 2.475\\
 TRADES & 1.648 & 2.167 &  1.751& 2.340\\ 
MART  & 1.603 & 2.409 &  1.732 & 2.421\\
\midrule
 I-ARD & 1.802 & 2.190&  1.925 & 2.110\\
 I-RSLAD & 1.808& 2.195&  1.888 & 2.051\\
 I-AdaAD  & 1.827& 2.114&  1.842& 1.950 \\
\midrule

 R-LwF & 1.553& 2.003&  1.731& 2.256\\
 R-LwF-mc & 0.994& 1.204&  0.895& 1.052\\
R-EWC-on  & 1.834& 2.841&  1.745& 2.629\\
 R-SI  & 1.578 & 2.119&  1.773& 2.342\\
\midrule
\textbf{FLAIR} w/o FPD  & 0.964& 1.139& 0.561&0.616\\
\textbf{FLAIR} & \underline{0.582}& \underline{0.665}&  \underline{0.492} & \underline{0.539} 
\\
 \textbf{FLAIR+} & \textbf{0.557}& \textbf{0.657}& \textbf{0.423} & \textbf{0.463} \\
 
 \bottomrule
 \end{tabularx}
   \caption{ Gradient Forgetting (GF), and Hessian Forgetting (HF). `w/o FPD' indicates without the FPD method.}
 \label{tab:GFHF}
\end{table}
\begin{figure}[h]
  \centering
  \includegraphics[width=0.75\columnwidth]{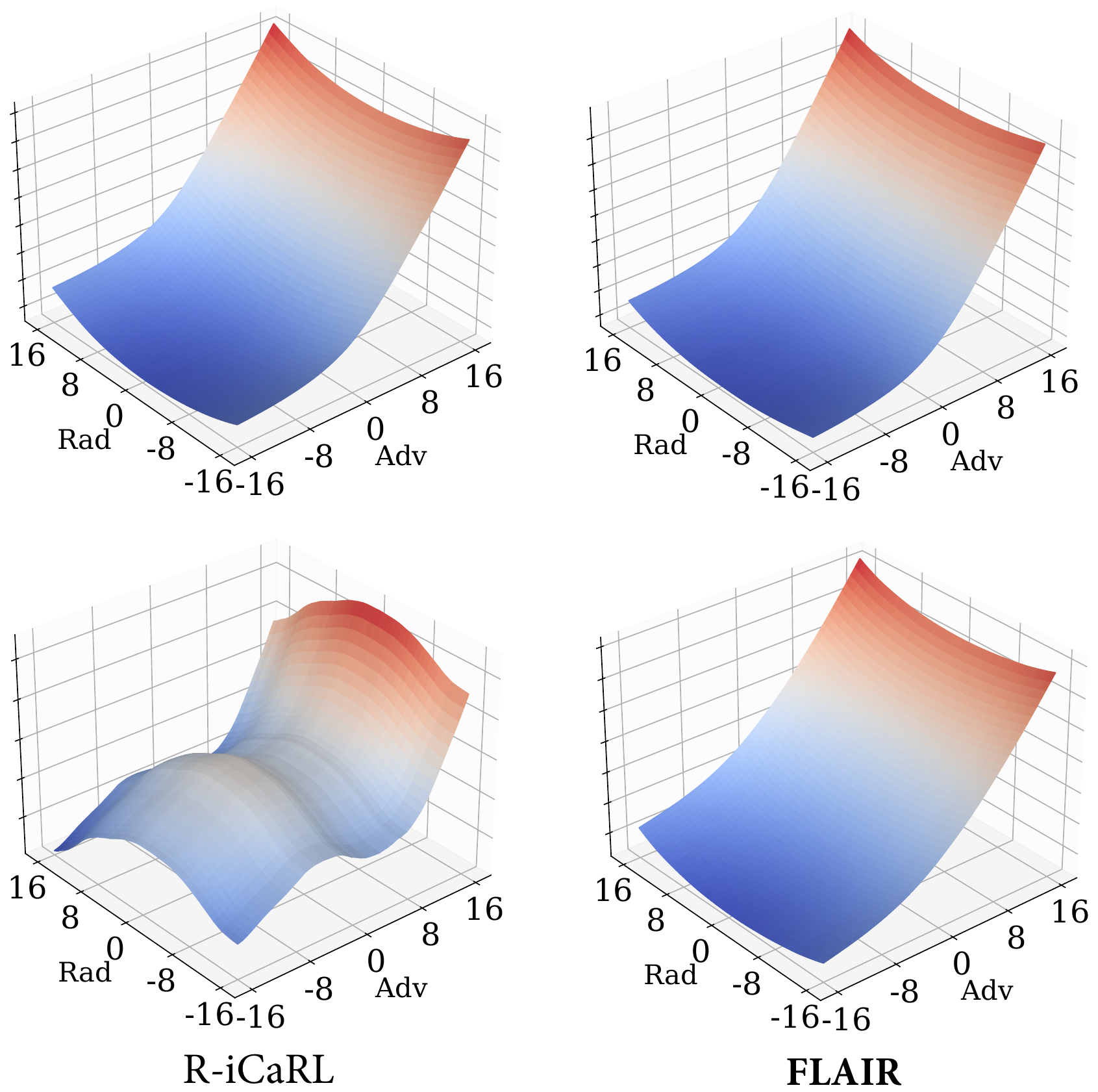}
  \caption{Loss landscape for each method at the beginning task $t$ (top row) and after learning task $t$ (bottom row).}
    \label{fig:visualization_loss}
\end{figure}

We measure gradient and Hessian forgetting, as defined in  \Cref{eqn:GFHF} to validate that FPD effectively preserves flatness.
In the \Cref{tab:GFHF}, note that all baselines exhibit the flatness forgetting problem, even including FLAIR without FPD.
This indicates that adversarial distillation alone is insufficient to address the flatness forgetting problem, resulting in a lack of robustness. 
On the other hand, applying FPD achieves the smallest GF and HF, demonstrating its effectiveness in preserving flatness and leading to strong robustness.
In \Cref{fig:visualization_loss}, we assess the loss landscape for flatness using the visualization approach \cite{park2021reliably}.
We plot cross-entropy loss projected in two directions: adversarial and random direction.
Comparing our method with the reasonably performing baseline R-iCaRL, we find that our approach effectively preserves flatness.

\begin{figure*}[t]
\centering 
    \includegraphics[width=1\textwidth] {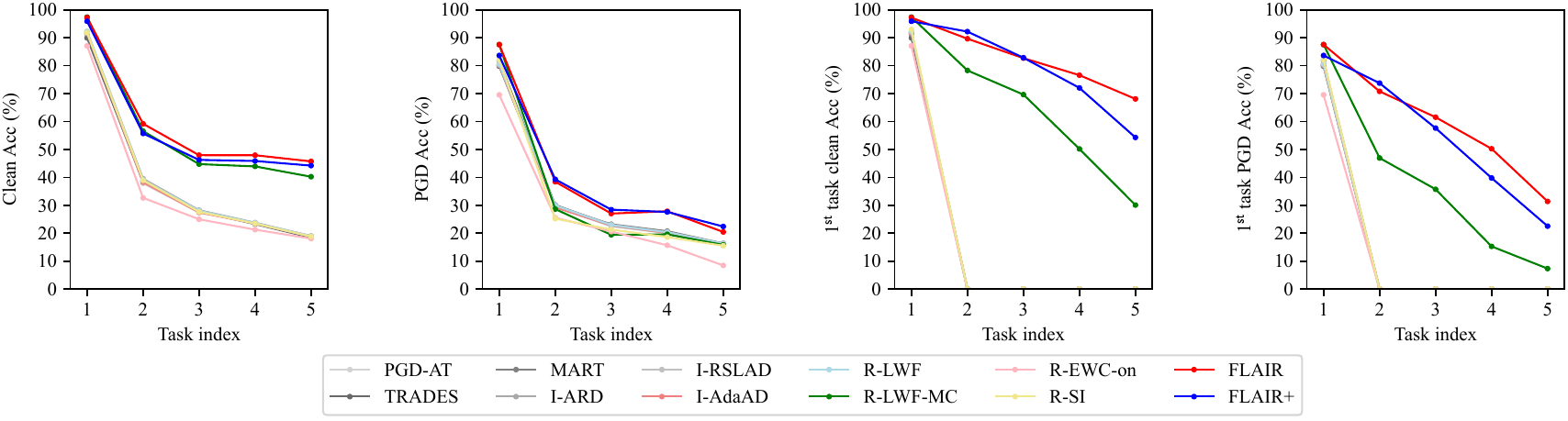}%
    \caption{
    Clean \textit{(Left)} and  20-step PGD \textit{(Middle left)} accuracy on all test datasets from the first to the current task, and clean \textit{(Middle right)} and 20-step PGD \textit{(Right)} accuracy on the 1$^\text{st}$ task's test dataset. Results are from various baselines and our methods on S-CIFAR100 across incremental task steps.
    }
    \label{fig:acc_by_task_step}
\end{figure*}

\subsection{Ablation Studies}
\textit{More detailed experimental settings and additional results are available in the appendix.}

\subsubsection{Experimental results after each task}
In \Cref{fig:acc_by_task_step}, we illustrate the clean and PGD accuracy across incremental task steps for both the overall test dataset and the 1$^\text{st}$ task's test dataset. 
Our methods show strong performance with minimal forgetting.
Notably, except R-EWC-on and our methods, all other baselines exhibit a complete drop in clean and PGD accuracy on the 1$^\text{st}$ task's test dataset after the 1$^\text{st}$ task, indicating full forgetting of previous knowledge. 
This suggests that most baselines fail to retain knowledge from previous tasks when learning new ones, which leads to nearly ideal classification performance only on the most recent task's training dataset. 
Consequently, their clean and robust accuracy approach $100/T (\%)$, as shown in \Cref{tab:main_res18_nobuffer}.

\subsubsection{Sensitivity of hyperparameters}
\begin{figure}[hb]
  \centering
  \includegraphics[width=1\columnwidth]{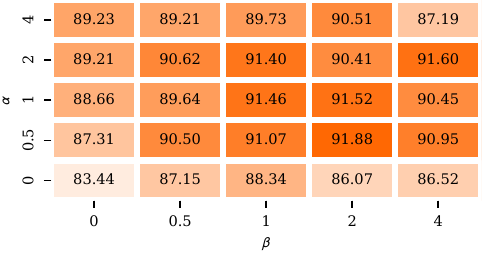}  
  \caption{Sensitivity of hyperparameter on S-CIFAR10 with 2000 size of memory buffer.}
  \label{fig:sensitivity}
\end{figure}

In \Cref{fig:sensitivity}, we measured hyperparameter sensitivity using the sum of clean and AutoAttack accuracy.
Generally, we see that performance improves as $\alpha$ and $\beta$ increase, with high performance within the range of $\alpha \in [0.5,1] $ and $\beta \in [1, 2]$.
we can see that the sensitivity does not appear to be significant.
Therefore, we selected the hyperparameters through a grid search, and the corresponding values are provided in the appendix.

\subsubsection{Effects of each component of FLAIR}
In \Cref{tab:component}, we investigated the effects of three core techniques in FLAIR, namely Adversarial Distillation with Separated Logit (ADSL), Flatness Preserving Distillation (FPD), and Augmentation (AUG).
As shown in the \Cref{tab:component}, each component contributes to performance improvement compared to when it is omitted.
Both ADSL and FPD reduce forgetting, while augmentation boosts the robustness.
Overall, the best results are achieved when all techniques are employed together. 
\begin{table}[h]
 \centering
 \setlength{\tabcolsep}{5.43pt}
\fontsize{9pt}{10pt}\selectfont
 \label{tab:component_detail}
 \begin{tabularx}{\columnwidth}{lll|cccc}
 \toprule
    ADSL & FPD  &AUG& Clean$\uparrow$ &  PGD$\uparrow$ & AA $\uparrow$ &  {R-BWT$\uparrow$} \\
    \midrule
   $\triangle$  &     && 22.55  & 11.07 & 9.23 &  -53.34  \\
     \midrule
   \checkmark &    && 40.24  & 16.60  & 15.84 &  -28.41 \\
    & \checkmark  &&  41.16 & 17.57  & 16.24 & -31.21 \\
 & & \checkmark  & 19.02& 16.42& 16.38&-72.00\\
\midrule
 \checkmark   & & \checkmark   & 40.23& 20.10& 18.07&-33.79\\
   & \checkmark   &\checkmark   & 43.92& 21.94&  19.66&  -25.21\\

 \checkmark & \checkmark   & & 43.64 & 18.42  & 17.02 &\textbf{-22.09} \\
 \midrule
 \checkmark   & \checkmark   & \checkmark   & \textbf{44.22}& \textbf{22.42}& \textbf{ 20.46}&-24.66\\
     \bottomrule
 \end{tabularx}
 \caption{Impact of each component of FLAIR on S-CIFAR10 without a memory buffer.
 The symbol $\triangle$ indicates ADSL with $\alpha = 0$.}
 \label{tab:component}
\end{table}

\section{Conclusion}

We have introduced FLAIR, a novel method designed to tackle the ARCIL problem. 
Our analysis identifies three major challenges within ARCIL: (1) the conflict between learning new tasks and retaining knowledge from previous tasks through distillation, (2) the disappearance of loss flatness, and (3) insufficient training data. 
To address these issues, FLAIR employs separated logits to reduce conflicts between new and old task knowledge, distills output differences from previous models to maintain loss flatness, and incorporates data augmentation to counteract the lack of training data.
This comprehensive strategy proves highly effective, establishing FLAIR as a strong baseline in this field.

\bibliography{aaai25}

\newpage

\appendix
\section*{Appendix}
\section{Comparision with TABA}
To the best of our knowledge, TABA \cite{bai2023towards} is the only work addressing the intersection of incremental learning and adversarial robustness, specifically in the rehearsal setting
The authors begin by employing a distillation technique similar to one of our baselines, R-iCaRL. 
They then introduce Task-Aware Boundary Augmentation (TABA) as an enhanced rehearsal method within the ARCIL framework.
Specifically, drawing on the insight that decision boundary examples are more susceptible to attacks and thus contribute more to adversarial robustness, they select two sets of boundary data—one from the rehearsal buffer and the other from the current dataset. These sets are augmented using Mixup to improve overall rehearsal performance.
However, their experiments are limited to S-CIFAR10 and CIFAR-100 (distinct from S-CIFAR100, as it comprises only five tasks with 20 classes each) on ResNet-18, with insufficient baselines for comprehensive evaluation.
Even their experimental results demonstrate insufficient robustness against AutoAttack, as shown in \Cref{tab:taba_paper}.
\begin{table}[h]
 \centering
 \setlength{\tabcolsep}{5.65pt}
 \begin{tabularx}{\linewidth}{lcccc}
 \toprule
  \multicolumn{1}{c}{\multirow{2}{*}{Method}} & \multicolumn{2}{c}{Result from} & \multicolumn{2}{c}{Accuracy (\%)}\\
  & TABA paper& Ours& Clean & AA \\

 \midrule
 
  R-iCaRL & \checkmark &  & 60.36 & 16.71\\
  R-iCaRL& & \checkmark & 57.78 & 25.87 \\
  \midrule
  TABA&  \checkmark & & 65.97 & 19.74\\
  TABA ($\lambda = 0.5$)&  & \checkmark & 40.48 & 18.45\\
  TABA ($\lambda = 0$)&  & \checkmark & 59.94& 24.18\\
  TABA+ ($\lambda = 0$)&  & \checkmark & 59.98 & 24.97\\
  \midrule
   \textbf{FLAIR}& & \checkmark & 63.81& 27.65\\
 \textbf{FLAIR+}& & \checkmark & 61.29& 30.06\\
\bottomrule
 \end{tabularx}
 \caption{Comparison between TABA paper results and our implementation results on S-CIFAR10 with a rehearsal buffer size of 2000 on ResNet-18.}
 \label{tab:taba_paper}
\end{table}

More concerning, they do not provide sufficient information for reproduction, including the absence of code and full training details. As a result, our implementation of TABA may yield different outcomes from those reported in the original paper. In our implementation, we observed that their primary contribution, boundary augmentation, negatively impacts overall performance. Contrary to their claim that the TABA hyperparameter $\lambda$ lies within $[0.45, 0.55]$, we found the optimal value to be close to 0 as shown in \Cref{tab:taba_paper}, suggesting no significant effect from the TABA method.
In contrast, our augmentation method enhances TABA's robustness, resulting in TABA+, as shown in \Cref{tab:taba_paper}.

TABA represents an important first step in addressing the rehearsal based ARCIL, offering a straightforward approach to solving it. However, the paper could benefit from more comprehensive experiments and baseline comparisons to fully validate their method, and the limited details for reproducibility pose some challenges.
Building on this foundation, we assert that our work is the first to thoroughly investigate the problem, baselines, methods, and experiments within the ARCIL framework.

\section{Baselines}

\begin{table*}[ht]
  \centering
 \setlength{\tabcolsep}{2.9pt}
    \fontsize{9pt}{10pt}\selectfont
    \begin{tabularx}{\linewidth}{c|l|l}
    \toprule
    \multicolumn{1}{c|}{\textbf{Type}} & \multicolumn{1}{c|}{\textbf{Methods}} & \multicolumn{1}{c}{\textbf{Formulation}} \\
    \midrule
\multirow{4}{*}{AT} & PGD-AT & $\mathbb{E}_{(\boldsymbol{x}, y)\sim \mathcal{S}_t} \big[l_{CE}(f_t(\boldsymbol{x}_{adv}),y)\big] $\\
\cmidrule{2-3}          & TRADES & $\mathbb{E}_{(\boldsymbol{x}, y)\sim \mathcal{S}_t} \big[l_{CE}(f_t(\boldsymbol{x}),y) + \alpha \cdot l_{KL}(f_t(\boldsymbol{x}_{adv}) \| f_t(\boldsymbol{x}))\big]  $\\
\cmidrule{2-3}          & MART & $\mathbb{E}_{(\boldsymbol{x}, y)\sim \mathcal{S}_t} \big[l_{BCE}(f_t(\boldsymbol{x}_{adv}),\boldsymbol{1}_y) + \alpha \cdot (1- \text{Pr}(f_t(x) = y)) \cdot l_{KL}(f_t(\boldsymbol{x}_{adv}) \| f_t(\boldsymbol{x}))\big]  $\\
 \midrule
\multirow{4}{*}{AD} & I-ARD & $\mathbb{E}_{(\boldsymbol{x}, y)\sim \mathcal{S}_t} \big[l_{CE}(f_t(\boldsymbol{x}_{adv}),y) + \beta \cdot l_{KL}([f_t(\boldsymbol{x}_{adv})]^{t-1}_{0} \| f_{t-1}(\boldsymbol{x}))\big] $\\
\cmidrule{2-3}          & I-RSLAD & $\mathbb{E}_{(\boldsymbol{x}, y)\sim \mathcal{S}_t} \big[l_{CE}(f_t(\boldsymbol{x}_{adv}),y) + \beta \!\cdot\! \big(\alpha \!\cdot\! l_{KL}([f_t(\boldsymbol{x}_{adv})]^{t-1}_{0} \| f_{t-1}(\boldsymbol{x})) + (1-\alpha) \!\cdot\! l_{KL}([f_t(\boldsymbol{x})]^{t-1}_{0} \| f_{t-1}(\boldsymbol{x}))\big)\big] $\\
\cmidrule{2-3}          & I-AdaAD & $\mathbb{E}_{(\boldsymbol{x}, y)\sim \mathcal{S}_t} \big[l_{CE}(f_t(\boldsymbol{x}_{adv}),y) + \beta \!\cdot\! \big( \alpha \!\cdot\! l_{KL}([f_t(\boldsymbol{x}_{adv})]^{t-1}_{0} \| f_{t-1}(\boldsymbol{x}_{adv})) + (1-\alpha) \!\cdot\! l_{KL}([f_t(\boldsymbol{x})]^{t-1}_{0} \| f_{t-1}(\boldsymbol{x}))\big)\big]$\\
 \midrule
    \multirow{6}{*}{\shortstack{Non-\\Rehearsal\\R-CIL}} &  R-EWC-on & $\mathbb{E}_{(\boldsymbol{x}, y)\sim \mathit{D_t}}\big[l_{CE}(f_t(\boldsymbol{x}_{adv}),y) + l_{EWC}(\theta_t, \theta_{t-1})\big]$   \\
    \cmidrule{2-3}    & R-LwF & $\mathbb{E}_{(\boldsymbol{x}, y)\sim \mathit{D_t}}\big[l_{CE}(f_t(\boldsymbol{x}_{adv}),y) + \alpha \cdot l_{KL}([f_t(\boldsymbol{x})]^{t-1}_{0} \| f_{t-1}(\boldsymbol{x}))\big]  $ \\
\cmidrule{2-3}          & R-LwF-MC & $\mathbb{E}_{(\boldsymbol{x}, y)\sim \mathit{D_t}}\big[l_{BCE} ([f_t(\boldsymbol{x}_{adv})]^{t}_{t-1},\boldsymbol{1}_y) \ +  l_{BCE}([f_{t}(\boldsymbol{x})]^{t-1}_{0}, {f}_{t-1}(\boldsymbol{x}))\big]$   \\
\cmidrule{2-3}& R-SI & $\mathbb{E}_{(\boldsymbol{x}, y)\sim \mathit{D_t}}\big[l_{CE}(f_t(\boldsymbol{x}_{adv}),y) + l_{SI}(\theta_t, \theta_{t-1})\big] $   \\
\midrule  
\multirow{9}{*}{\shortstack{Rehearsal\\R-CIL}}
& R-ER & $  \mathbb{E}_{(\boldsymbol{x}, y)\sim \mathit{D_t}}\big[l_{CE}(f_t(\boldsymbol{x}_{adv}),y)\big] + \mathbb{E}_{(\boldsymbol{x}, y)\sim \mathit{M}} \big[l_{CE}(f_t(\boldsymbol{x}_{adv}),y)\big]  $   \\
\cmidrule{2-3}          & R-ER-ACE & $  \mathbb{E}_{(\boldsymbol{x}, y)\sim \mathit{D_t}}\big[l_{ACE}(f_t(\boldsymbol{x}_{adv}),C_{curr})\big] + \mathbb{E}_{(\boldsymbol{x}, y)\sim \mathit{M}} \big[l_{CE}(f_t(\boldsymbol{x}_{adv}),y)\big] $   \\
\cmidrule{2-3}          & R-DER & $  \mathbb{E}_{(\boldsymbol{x}, y)\sim \mathit{D_t}}\big[l_{CE}(f_t(\boldsymbol{x}_{adv}),y)\big] + \mathbb{E}_{(\boldsymbol{x}, \mathbf{z})\sim \mathit{M}} \big[ \alpha \cdot 
 l_{MSE}(f_t(\boldsymbol{x}_{adv}),\mathbf{z})\big] $ \\
\cmidrule{2-3}          & R-DER++ & $  \mathbb{E}_{(\boldsymbol{x}, y)\sim \mathit{D_t}}\big[l_{CE}(f_t(\boldsymbol{x}_{adv}),y)\big] + \mathbb{E}_{(\boldsymbol{x}, \mathbf{z}, y) \sim \mathit{M}} \big[\alpha \cdot l_{MSE}(f_t(\boldsymbol{x}_{adv}),\mathbf{z}) + \beta \cdot l_{CE}(f_t(\boldsymbol{x}_{adv}),y)\big] $\\
\cmidrule{2-3}          & R-iCaRL & $\mathbb{E}_{(\boldsymbol{x}, y)\sim \mathit{D_t} \cup B_{t-1}}\big[l_{BCE} ([f_t(\boldsymbol{x}_{adv})]^{t}_{t-1},\boldsymbol{1}_y) \ +  l_{BCE}([f_{t}(\boldsymbol{x})]^{t-1}_{0}, {f}_{t-1}(\boldsymbol{x}))\big]$ \\
\cmidrule{2-3}          & R-LUCIR   &  $\mathbb{E}_{(\boldsymbol{x}, y)\sim \mathit{D_t} \cup B_{t-1}}\big[l_{CE} (f_t(\boldsymbol{x}_{adv}),y) + \alpha \cdot l_{dis}^G(\boldsymbol{x}_{adv}) \big] \ + \mathbb{E}_{(\boldsymbol{x}, y)\sim B_{t-1}}\big[ \beta \cdot l_{mr}({\boldsymbol{x}_{adv}})\big]$ \\

\midrule
\multirow{2.5}{*}{ARCIL} & TABA & $\mathbb{E}_{(\boldsymbol{x}, y)\sim \mathit{D_t} \cup B_{t-1} \cup A_\text{TABA}}\big[l_{BCE} ([f_t(\boldsymbol{x}_{adv})]^{t}_{t-1},\boldsymbol{1}_y) \ +  l_{BCE}([f_{t}(\boldsymbol{x}_{adv})]^{t-1}_{0}, {f}_{t-1}(\boldsymbol{x}))\big]$ \\
\cmidrule{2-3}          & \textbf{FLAIR} & $\mathbb{E}_{(\boldsymbol{x}, y)\sim S_t}\big[l_{BCE} ([f_t(\boldsymbol{x}_{adv})]^{t}_{t-1},\boldsymbol{1}_y) + \alpha \cdot l_{BCE}([f_{t}(\boldsymbol{x}_{adv})]^{t-1}_{0}, {f}_{t-1}(\boldsymbol{x}_{adv}))  
+ \beta \cdot l_{FPD}(\boldsymbol{x}, \boldsymbol{x}_{adv} ; f_t, f_{t-1})\big]$\\
    \midrule
    \multicolumn{3}{p{17.55cm}}{
    \textbf{Notation}
    $f_t(\cdot)$ is $t$-th task training model, while $f_{t-1}(\cdot)$ is the $(t-1)$-th task learned model.
    $\boldsymbol{x}_{adv}$ is the adversarial image of $\boldsymbol{x}$.
    The notation $[\cdot]_i^j$ represents a slicing operation that returns the outputs immediately after the $i$-th task up to and including the $j$-th task output, and $\boldsymbol{1}_y$ is a one-hot encoding for the label $y$.
    $\mathit{D_t}$ is the current task dataset, and $B_{t-1}$ is buffered data from past dataset after training the $(t-1)$-th task, and $M$ is an online-updated memory buffer, and $\mathbf{z}$ is buffered logits.
    $S_t$ represents the current training dataset for methods applicable in both rehearsal and non-rehearsal settings, instead of $D_t$ or $D_t \cup B_{t-1}$.
    $l_{CE}$, $l_{KL}$, $l_{BCE}$, and $l_{MSE}$ are cross-entropy, KL divergence, binary cross-entropy, and mean-squared error losses, respectively.
    $\alpha$, and $\beta$ are the hyperparameters.
    Both $l_{EWC}$ and $l_{SI}$ refer to regularization loss for model parameters. 
    $l_{ACE}(\cdot, C_{curr})$ is a modified cross-entropy loss that constrains the class set used in the denominator to the classes represented in the incoming batch.
    $l_{dis}^G(\cdot)$ and $l_{mr}(\cdot)$ are distillation losses on the features and margin ranking loss of the original LUCIR method.
    $A_\text{TABA}$ represents Task-Aware Boundary Augmentation proposed in TABA.}\\
    \bottomrule
    \end{tabularx}

      \caption{Different methods to fit in ARCIL setting. Type AT stands for naive Adversarial Training on ARCIL, while AD is revised adversarial distillation methods by considering the previous task model as a teacher model.
R-CIL consists of a set of revised CIL methods, mainly by changing the input $\boldsymbol{x}$ of the learning incremented task to $\boldsymbol{x}_{adv}$. R-CIL is further categorized into two subtypes: Non-Rehearsal and Rehearsal types.}
  \label{tab:cil_adv_frame_supp}
\end{table*}

we construct new baselines for ARCIL by applying adversarial training on those incremental learning in \Cref{tab:cil_adv_frame_supp}.

\subsection{PGD-AT} \citeauthor{PGD} propose PGD-AT (Projected Gradient Descent Adversarial Training) to improve robustness by integrating adversarial examples into training. This method generates adversarial examples using the PGD algorithm, which perturbs inputs to maximize the model's loss within a specified norm constraint. In the ARCIL framework, PGD-AT is applied as a standard adversarial training technique and does not incorporate methods specifically designed for incremental learning.

\subsection{TRADES} \citeauthor{TRADES} introduce TRADES  to enhance model robustness by balancing the trade-off between adversarial robustness and clean accuracy. This approach generates adversarial examples and applies a loss function that combines both cross-entropy loss on clean examples and a KL divergence term that measures the discrepancy between the model’s predictions on clean and adversarial inputs. 
In the ARCIL framework, TRADES is used as a conventional adversarial training method and does not integrate strategies specifically tailored for incremental learning.

\subsection{MART} \citeauthor{MART} propose MART (Misclassification Aware adveRsarial Training) to improve robustness by incorporating adversarial examples during training while adjusting the loss function to introduce per-sample weights based on the confidence of each sample. 
MART uses a combination of binary cross-entropy loss on adversarial examples and a term that emphasizes the model’s uncertainty, based on its prediction probabilities. 
In the ARCIL framework, MART is utilized as a standard adversarial training technique and does not include methods specifically designed for incremental learning.

\subsection{I-ARD, I-RSLAD, and I-AdaAD} 
We adapted the adversarial distillation methods proposed by \citeauthor{ard, rslad, adaad}—specifically, I-ARD, I-RSLAD, and I-AdaAD—to align with the ARCIL framework. These methods extend the concept of knowledge distillation from class-incremental learning to adversarial settings, thereby enhancing model robustness while preserving knowledge from previous tasks. 
Although their loss functions slightly differ, all three methods aim to preserver past adversarial robustness through adversarial distillation.

\subsection{R-EWC-on}
We adapted the Elastic Weight Consolidation (EWC) method, originally proposed by \citeauthor{ewc} to address catastrophic forgetting in class-incremental learning, for the ARCIL framework. EWC introduces a regularization term in the loss function that penalizes significant changes to model parameters important for previous tasks. In the ARCIL context, this method is applied to adversarially perturbed examples, helping the model preserve robustness from earlier tasks while learning new ones incrementally.

\subsection{R-LwF}
We adapted the Learning without Forgetting (LwF) method proposed by \citeauthor{lwf} to the ARCIL framework, resulting in R-LwF. LwF mitigates catastrophic forgetting by using knowledge distillation to maintain performance on previous tasks while learning new ones. In the ARCIL context, R-LwF incorporates adversarially perturbed examples into the cross-entropy loss for current tasks, ensuring that the model learns robustly from adversarial inputs. Simultaneously, it includes a knowledge distillation term that preserves performance on past tasks, using outputs from previous models as targets, thereby balancing the model’s robustness and incremental learning.

\subsection{R-LwF-MC}
We adapted the Learning without Forgetting applied to multi-class classification (LwF-MC) method from \citeauthor{icarl} to fit the ARCIL framework, resulting in R-LwF-MC. This approach is similar to the Learning without Forgetting (LwF) technique but reflects the original iCaRL loss when no memory buffer is used. R-LwF-MC utilizes adversarially perturbed examples and applies binary cross-entropy loss to compare the model's outputs on both current and past tasks, facilitating knowledge retention from previous tasks while learning new ones incrementally.

\subsection{R-SI}
We adapted the Synaptic Intelligence (SI) method proposed by \citeauthor{synaptic} to fit the ARCIL framework, resulting in R-SI. SI, originally developed to address catastrophic forgetting in class-incremental learning, introduces a regularization term that penalizes significant changes to model parameters important for previous tasks. In R-SI, this approach is applied to adversarially perturbed examples, using cross-entropy loss for current tasks and the SI regularization to retain robustness from previous tasks while incrementally learning new ones.

\subsection{R-ER}
We adapted the basic Experience Replay (ER) method proposed by \citeauthor{er} to the ARCIL framework. Originally developed for class-incremental learning (CIL), ER is a fundamental technique that maintains a memory buffer of past examples to mitigate catastrophic forgetting during training. In the ARCIL setting, this approach is employed to replay adversarially perturbed examples, helping the model retain robustness across incremental tasks.

\subsection{R-ER-ACE}
We adapted the Experience Replay with Asymmetric Cross-Entropy (ER-ACE) method to the ARCIL framework, building on the foundational work by \citeauthor{er_ace}. ER-ACE extends the basic ER approach by introducing a modified cross-entropy loss, $l_{ACE}(\cdot, C_{curr})$, which constrains the class set in the loss function to the classes represented in the incoming batch. In the ARCIL setting, this method is used to replay adversarially perturbed examples while adapting the loss to ensure effective class-specific learning, thus helping the model maintain robustness and accuracy across incremental tasks.

\subsection{R-DER and R-DER++}
We adapted the Dark Experience Replay (DER) method from \citeauthor{der} for the ARCIL framework, resulting in R-DER. Originally designed for class-incremental learning, DER uses a memory buffer and dark knowledge distillation to address catastrophic forgetting. In ARCIL, R-DER applies adversarially perturbed examples and employs a cross-entropy loss for current tasks alongside a mean-squared error loss on buffered logits. R-DER++ builds on this by incorporating cross-entropy and mean-squared error losses, enhancing the model's ability to effectively manage adversarial robustness and incremental learning.

\subsection{R-iCaRL}
We adapted the Incremental Classifier and Representation Learning (iCaRL) method proposed by \citeauthor{icarl} to fit the ARCIL framework, resulting in R-iCaRL. This method uses adversarially perturbed examples and applies binary cross-entropy loss to the current task's adversarial inputs. For distillation, R-iCaRL uses binary cross-entropy loss to compare the outputs of the model on past tasks. This approach ensures that the model maintains robustness to adversarial attacks while retaining knowledge from previously learned tasks.

\subsection{R-LUCIR}
We adapted the Learning a Unified Classifier Incrementally via Rebalancing (LUCIR) method proposed by \citeauthor{lucir} to fit the ARCIL framework, resulting in R-LUCIR. This approach uses adversarially perturbed examples and applies cross-entropy loss to these examples for the current task. The method also incorporates distillation losses, where the feature distillation loss is computed using adversarially perturbed examples, while the margin ranking loss is applied separately. This adaptation helps maintain robustness to adversarial attacks while learning new tasks incrementally, ensuring effective knowledge retention and adaptation.

\subsection{TABA}
TABA \cite{bai2023towards} is the only work that combines incremental learning with adversarial robustness in rehearsal settings. The study begins by using a distillation method akin to baseline R-iCaRL, then introduces Task-Aware Boundary Augmentation (TABA) to enhance the rehearsal process within the ARCIL framework. TABA focuses on boundary examples, which are more vulnerable to adversarial attacks, by selecting and augmenting these examples from both the rehearsal buffer and the current dataset using Mixup.

\begin{table*}[ht]
 \centering
 \setlength{\tabcolsep}{6.1pt}
    \fontsize{9pt}{10pt}\selectfont
 \begin{tabular}{lrrrrrrrrrrrr}

 \toprule

   \multicolumn{1}{c}{\multirow{2}{*}{Method}} & \multicolumn{4}{c}{S-CIFAR10}& \multicolumn{4}{c}{S-CIFAR100} & \multicolumn{4}{c}{S-SVHN}\\
 \cmidrule(l){2-5}
\cmidrule(l){6-9}
 \cmidrule(l){10-13}
   & Clean$\uparrow$ & PGD$\uparrow$ & AA$\uparrow$ & R-BWT$\uparrow$& Clean$\uparrow$ & PGD$\uparrow$ & AA$\uparrow$ & R-BWT$\uparrow$  & Clean$\uparrow$ & PGD$\uparrow$ & AA$\uparrow$ &R-BWT$\uparrow$  \\
\midrule

 PGD-AT & 34.71& 19.80& 19.30&  -65.81& 12.06 & 5.59 & 5.50 & -35.54 & 41.03& 14.30& 9.91&-70.13\\
  TRADES & 23.00& 17.03& 16.92&  -69.88& 11.31 & 6.30 & 6.21 & -44.06  & 35.91& 15.51& 10.93&-71.90\\
  MART & 25.51& 18.69& 18.25&  -68.45& 11.36& 5.62& 5.49&   -36.27& 42.64& 15.73& 10.91&-69.25\\
\midrule

 I-ARD & 28.21& 19.50& 19.24&  -68.65& 9.24& 5.69& 5.40&   -44.88& 31.12& 12.05& 7.79&-69.06\\
  I-RSLAD& 28.46& 19.87& 19.68&  -68.48& 8.75& 5.65& 5.42&   -45.71& 32.16& 12.43& 8.16&-68.45\\
  I-AdaAD& 27.82& 19.46& 19.24&  -68.45& 8.35& 5.27& 4.99&   -43.72& 31.24& 12.65& 7.96&-67.24\\
\midrule

 R-ER & 32.77& 17.99& 17.71& -64.45& 10.95 & 5.43 & 5.27 & -35.58 & 35.61& 13.13& 8.94&-70.15\\

  R-ER-ACE & 52.42& 13.55& 13.12& -30.30& 22.11 & 3.84 & 3.53 & -17.48 & 67.58& 18.27& 19.07&-27.95\\
  R-DER & 23.68& 16.33& 16.18& -65.74& 12.93 & 5.85 & 5.15 & -37.14 & 31.44& 11.82& 8.14&-69.88\\ 
  R-DER++ & 24.63& 16.24& 16.09& -67.51& 12.61 & 5.71 & 5.21 & -34.01 & 40.12& 13.24& 9.76&-65.29\\ 
  R-iCaRL & \underline{52.53}& 24.41& 23.26& -20.68& 32.14& 8.58& 6.86&  -20.50& 65.29& 11.06& 9.45&-46.08\\ 
  R-LUCIR & 48.13& 22.67& 21.88& -65.21& 17.74& 7.31& 6.85&  -36.27& 42.11& 17.13& 12.19&-64.87\\
\midrule
 TABA & \textbf{53.83}& 22.10& 21.09&  -29.36& 16.88& 4.28& 3.79& -24.59& 49.12& 9.23& 5.79&-46.52\\

  \textbf{FLAIR}&  52.33& \underline{25.36}& \underline{24.11}&  \textbf{-17.05}& \underline{35.21}& \underline{13.26}& \underline{10.48}& \underline{-14.11}& \underline{69.22}& \underline{21.77}& \underline{22.45}&\underline{-12.37}\\
 \textbf{FLAIR+}& 50.80& \textbf{26.83}& \textbf{24.74}& \underline{-19.85}& \textbf{35.26}& \textbf{16.09}& \textbf{12.52}&  \textbf{-12.33}& \textbf{74.46}& \textbf{36.83}& \textbf{35.13}& \textbf{-8.51}\\
 \bottomrule
 \end{tabular}
     \caption{Clean accuracy, 20-step PGD accuracy, Autoattack (AA) performance (\%), and Robust Backward Transfer (R-BWT) measured on ResNet-18 for S-CIFAR10, S-CIFAR100, and S-SVHN with 500 size of memory buffer.}
 \label{tab:res18_buffer_500}
\end{table*}

\section{More Experimental Results}
In this section, We conduct further experiments to corroborate our main contribution.

\subsection{Different Buffer Size for Rehearsal Method}
In \Cref{tab:res18_buffer_500}, we present experiments evaluating the performance of different buffer sizes for the rehearsal method.
We use a buffer size of 500, in contrast to the 2000 used in the main paper, which results in overall lower performance compared to the results reported with the larger buffer in the paper. Despite this, our method still outperforms the baselines in terms of robustness across all datasets, achieving the highest clean accuracy on both S-CIFAR100 and s-SVHN.

\subsection{Randomness Check for FLAIR}
\begin{table}[ht]
 \centering
 \setlength{\tabcolsep}{6pt}
 \begin{tabular}{lccccc}
 \toprule
   Method &Round& Clean & PGD & AA & R-BWT$\uparrow$  \\
\midrule
  \multirow{6}{*}{\textbf{FLAIR}}& 1 & 43.75&18.46& 16.96& -21.36\\
   &2& 44.06&18.84&17.31&-22.58\\
   &3& 43.05& 18.36&  16.93& -22.94\\
 & 4& 43.70& 18.02& 16.88&-21.48\\
 & Mean& 43.64& 18.42& 17.02&-22.09\\
\midrule
  \multirow{6}{*}{\textbf{FLAIR+}}&1& 44.32& 22.00& 20.51& -24.10\\
   &2& 44.70& 21.85& 20.35&-25.51\\
   &3& 43.84& 22.92& 20.70& -25.14\\
 & 4& 44.02& 22.91& 20.28&-23.89\\
 & Mean& 44.22& 22.42& 20.46&-24.66\\
 \bottomrule
 \end{tabular}
    \caption{Adversarial training results on ResNet-18 with various methods on S-CIFAR10 without memory buffer.}
 \label{tab:randomness}
\end{table}
In \Cref{tab:randomness}, we evaluate the impact of randomness on our methods by varying the random seed while keeping the same settings and hyperparameters. 
The results reveal a small variance in performance, within 0.6 percentage points. 
In comparison, R-BWT exhibits a slightly higher variance, within 1 percentage point.
Given the ARCIL setup, where adversarial training occurs across several sequential task datasets, these reported variations are reasonable. 
Notably, even under the worst-case scenario, our method consistently outperforms the baselines.

\subsection{Experiments on S-TinyImageNet}
\begin{table}[ht]
 \centering
 \setlength{\tabcolsep}{9.35pt}
 \begin{tabularx}{\linewidth}{lrrrr}
 \toprule
   Method& Clean & PGD & AA & R-BWT$\uparrow$  \\
\midrule
 PGD-AT & 6.24 & 2.32 & 2.25 & -24.52\\
  TRADES & 6.09 & 2.63 & 2.47 & -30.77\\
  MART & 6.12 & 2.33 &  2.26 & -24.50 \\
\midrule

 I-ARD &  5.99& 3.16 & 2.51 & -31.70  \\
  I-RSLAD& 5.72& 3.25  &  2.51 & -30.30 \\
  I-AdaAD& 4.69  & 2.61  & 1.88 & -26.87\\
\midrule

 R-ER & 8.51 & 2.46 & 2.39 & -22.32  \\
 R-ER-ACE & 18.73 & 1.84  & 1.68 & -13.33   \\
  R-DER &  10.87 & 3.02  & 2.15  & -27.96\\
  R-DER++ & 11.29 & 2.79 & 2.27 & -22.01 \\
R-iCaRL & 19.30 & 5.10  & 3.48  & -21.86\\
R-LUCIR & 11.36 & 2.72  & 2.23 & -24.30\\
TABA & 19.70 &  5.51 &  3.58 & -21.78\\
\midrule
  \textbf{FLAIR}&  \textbf{28.11} & \underline{9.91} & \underline{6.49} &  \textbf{-6.88} \\
  \textbf{FLAIR+}& \underline{27.62} &  \textbf{11.39} &  \textbf{7.78}  &  \underline{-8.70}  \\
 \bottomrule
 \end{tabularx}
\caption{Adversarial training results on ResNet-18 with various methods on S-TinyImageNet with memory buffer 2,000.}
 \label{tab:res18_tinyimg_buffer_2000}
\end{table}
We conduct experiments on S-TinyImageNet using a memory buffer setting. Note that without the buffer, the overall performance is significantly reduced; therefore, we use the memory buffer setting for our experiments. In \Cref{tab:res18_tinyimg_buffer_2000}, our method demonstrates strong robustness and high clean accuracy with minimal forgetting.


\begin{table*}[t]
 \centering
 \setlength{\tabcolsep}{6.1pt}
    \fontsize{9pt}{10pt}\selectfont
 \begin{tabular}{lrrrrrrrrrrrr}

 \toprule

   \multicolumn{1}{c}{\multirow{2}{*}{Method}} & \multicolumn{4}{c}{S-CIFAR10}& \multicolumn{4}{c}{S-CIFAR100} & \multicolumn{4}{c}{S-SVHN}\\
 \cmidrule(l){2-5}
\cmidrule(l){6-9}
 \cmidrule(l){10-13}
   & Clean$\uparrow$ & PGD$\uparrow$ & AA$\uparrow$ & R-BWT$\uparrow$& Clean$\uparrow$ & PGD$\uparrow$ & AA$\uparrow$ & R-BWT$\uparrow$  & Clean$\uparrow$ & PGD$\uparrow$ & AA$\uparrow$ &R-BWT$\uparrow$  \\
\midrule
 PGD-AT &18.65 & 16.00 & 15.94& -71.78  & 8.38 & 5.71 & 5.57 & -46.13 & 17.55 & 10.69 & 6.38445 & -74.50\\
  TRADES & 18.13 & 15.74 & 15.72 & -71.51  & 8.12 & 5.76 & 5.53 & -46.98 & 19.05 & 10.97 & 6.66 & -75.83\\
  MART & 18.60 & 16.08& 16.06& -72.20 & 7.32 &  5.38& 4.99 & -40.22   &16.92 &12.02 & 7.30& -76.52\\
\midrule

 I-ARD & 39.06 & 25.88 & 25.42 & -58.78  & 7.61 & 4.80 & 4.33 &  -34.32 & 61.78 & 23.32 & 19.34& -54.22\\
  I-RSLAD& 39.15&26.23 & 25.72& -58.38  & 7.36 & 4.85 & 4.33 & -34.25  & 61.21  & 22.61 & 18.46 & -54.58\\
  I-AdaAD& 37.59&25.25 & 24.83&  -58.57 & 7.03 & 4.52 & 4.04 & -32.41  & 59.76 & 22.31 & 17.68 & -55.36 \\
\midrule

 R-ER & 51.14&23.07 &22.35&-53.60 & 16.12 & 6.46 & 6.27 & -28.86 & 57.70 & 19.18 & 15.77 & -53.55\\

  R-ER-ACE & 62.76 & 19.28 & 18.14 &  -24.01& 29.94 & 7.01 & 6.61 & -12.76 & 72.32 & 21.61 &21.35 & -19.36
\\

  R-DER & 20.49& 13.43&12.36& -67.61& 10.41 & 6.11 & 5.59 & -40.42 & 9.96 & 9.83 & 5.75 & -57.96\\ 
  R-DER++ & 
21.58&  15.00 & 14.72& -62.07 & 17.85 & 6.65 & 6.15 & -35.15 &28.97 &7.55 &4.45 &-54.49\\ 
  R-iCaRL & 55.44 & 28.70 & 26.41 & -17.20 & 27.44 & 13.19 & 10.13 & -10.94  & 71.99 & 20.62 & 17.29 &-32.40 \\ 
  R-LUCIR & 56.05 & 30.51 & 29.00 &-50.21 & 24.87 & 9.15 & 8.67 & -29.42  & 61.41 & 23.80 & 19.54 & -52.75\\
\midrule
 TABA & 54.08 &  20.82 & 19.73 & -18.57  & 12.12 &  3.27& 2.60& -14.85 & 69.34 & 10.69 &6.66 &-36.73\\
  \textbf{FLAIR}& \textbf{59.60} & \underline{30.29} & \underline{27.32} & \textbf{-10.00}  & \textbf{32.64} & \underline{15.08} & \underline{11.92} & \textbf{-7.95} & 77.61 & 29.22 & 26.72 & -7.13\\
 \textbf{FLAIR+}& \underline{56.82} & \textbf{31.07} & \textbf{28.07} & \underline{-10.51} & \underline{30.87} & \textbf{16.29} & \textbf{12.26} & \underline{-8.90} & 83.28&42.29 & 38.27&-3.32\\
 \bottomrule
 \end{tabular}
     \caption{Clean accuracy, 20-step PGD accuracy, Autoattack (AA) performance (\%), and Robust Backward Transfer (R-BWT) measured on MobileNet-v2 for S-CIFAR10, S-CIFAR100, and S-SVHN with 2000 size of memory buffer.}
 \label{tab:mnv2_buffer_2000}
\end{table*}

\begin{table*}[!ht]
 \centering
 \setlength{\tabcolsep}{6.1pt}
    \fontsize{9pt}{10pt}\selectfont
 \begin{tabular}{lrrrrrrrrrrrr}

 \toprule

   \multicolumn{1}{c}{\multirow{2}{*}{Method}} & \multicolumn{4}{c}{S-CIFAR10}& \multicolumn{4}{c}{S-CIFAR100} & \multicolumn{4}{c}{S-SVHN}\\
 \cmidrule(l){2-5}
\cmidrule(l){6-9}
 \cmidrule(l){10-13}
   & Clean$\uparrow$ & PGD$\uparrow$ & AA$\uparrow$ & R-BWT$\uparrow$& Clean$\uparrow$ & PGD$\uparrow$ & AA$\uparrow$ & R-BWT$\uparrow$  & Clean$\uparrow$ & PGD$\uparrow$ & AA$\uparrow$ &R-BWT$\uparrow$  \\
\midrule

 PGD-AT+ & 56.83 & 29.25 & 27.54 &  -54.50& 23.44 &9.19 & 8.74& -34.40& 71.00& 29.94& 25.30&-49.07\\
  TRADES+ & 49.26 & 25.48& 25.19 &  -62.66& 18.34& 9.88 & 9.34& -41.20& 48.14& 23.18& 20.19&-64.14\\
  MART+ & 40.20 & 25.78 & 23.84 &  -60.62 & 21.61& 9.97& 9.44&   -41.38& 61.70& 31.43& 25.30&-57.55\\
\midrule

 I-ARD+ & 40.84 & 27.75 & 27.30&  -56.75 & 13.90& 8.17& 7.59&   -39.83& 70.34& 33.51& 29.04&-52.44\\
  I-RSLAD+& 41.77 & 28.26 & 27.69  &  -56.11& 13.03& 7.82& 7.17&   -40.26& 70.48& 34.08& 29.40&-52.01\\
  I-AdaAD+& 40.55 & 27.56& 27.09 &  -56.52& 9.98& 6.04& 5.34&   -38.08& 66.26& 32.43& 27.11&-53.68\\
\midrule

 R-ER+ & 52.40 & 23.15 & 22.50 & -52.08 & 16.41& 6.77& 6.50& -34.10& 58.19& 22.03& 17.76&-60.93\\

  R-ER-ACE+ & 62.44 & 20.65  & 19.93  & -23.21 & 27.62& 6.00& 4.52& -16.78& 80.66& 29.36& 28.82&-24.54\\
  R-DER+ & 21.16 & 16.26 & 16.20 & -61.70 & 14.28& 6.83& 5.64& -43.69& 32.98& 11.24& 7.19&-71.12\\ 
  R-DER++/+ & 24.23 & 16.39& 16.21  & -61.32 & 19.91& 7.54& 6.71& -36.96& 20.50& 11.30& 6.81&-77.23\\ 
  R-iCaRL+ & 55.44 & 29.03& 26.55  & -10.09& 34.75& 17.29& 13.61 & -11.90  & 79.77& 33.46& 32.12&-16.24\\ 
  R-LUCIR+ & 61.54& 31.09& 29.81& -56.48& 29.02& 11.68& 10.66&  -35.46& 72.51& 31.08& 26.08&-53.11\\
\midrule
 TABA+ & 59.98 & 26.87& 24.97  &  -16.74& 33.04& 9.88& 8.30&  -17.71& 81.37& 35.50& 33.16&-19.93\\
 \textbf{FLAIR+} &61.29 &  \textbf{33.35} & \textbf{30.06} & \textbf{-12.49}& \textbf{38.04}& \textbf{18.03}& \textbf{14.43}&  \textbf{-8.43}& \textbf{83.83}& \textbf{44.33}& \textbf{40.10}& \textbf{-2.74}\\
 \bottomrule
 \end{tabular}
     \caption{Clean accuracy, 20-step PGD accuracy, Autoattack (AA) performance (\%), and Robust Backward Transfer (R-BWT) measured on ResNet-18 for S-CIFAR10, S-CIFAR100, and S-SVHN with 2000 size of memory buffer with data augmentation.}
 \label{tab:res18_buffer_2000_aug}
\end{table*}

\subsection{Experiments on another architecture}
We extended our experiments to other architectures, including MobileNet-V2, as shown in \Cref{tab:mnv2_buffer_2000}, and our approach our approach consistently demonstrated superior performance. Since our method relies solely on output-based losses, without utilizing intermediate features from the model, it can be applied effectively regardless of the architecture used.

\subsection{Experiments with Augmentation}
We further analyze whether the augmentation technique is effective across other baselines, as shown in \Cref{tab:res18_buffer_2000_aug}.
We observe that applying either RA or AuA techniques to most baseline methods results in a slight improvement in robustness compared to the original baseline methods. However, for certain methods like R-DER and R-DER++, the performance is even lower than the original methods.
This suggests augmentation alone is not always optimal. More importantly, FLAIR+, consistently outperforms the augmented baselines, demonstrating its effectiveness.

\subsection{Further Visualization of Loss Landscape}
\begin{figure*}[h]
  \centering
  \includegraphics[width=1\linewidth]{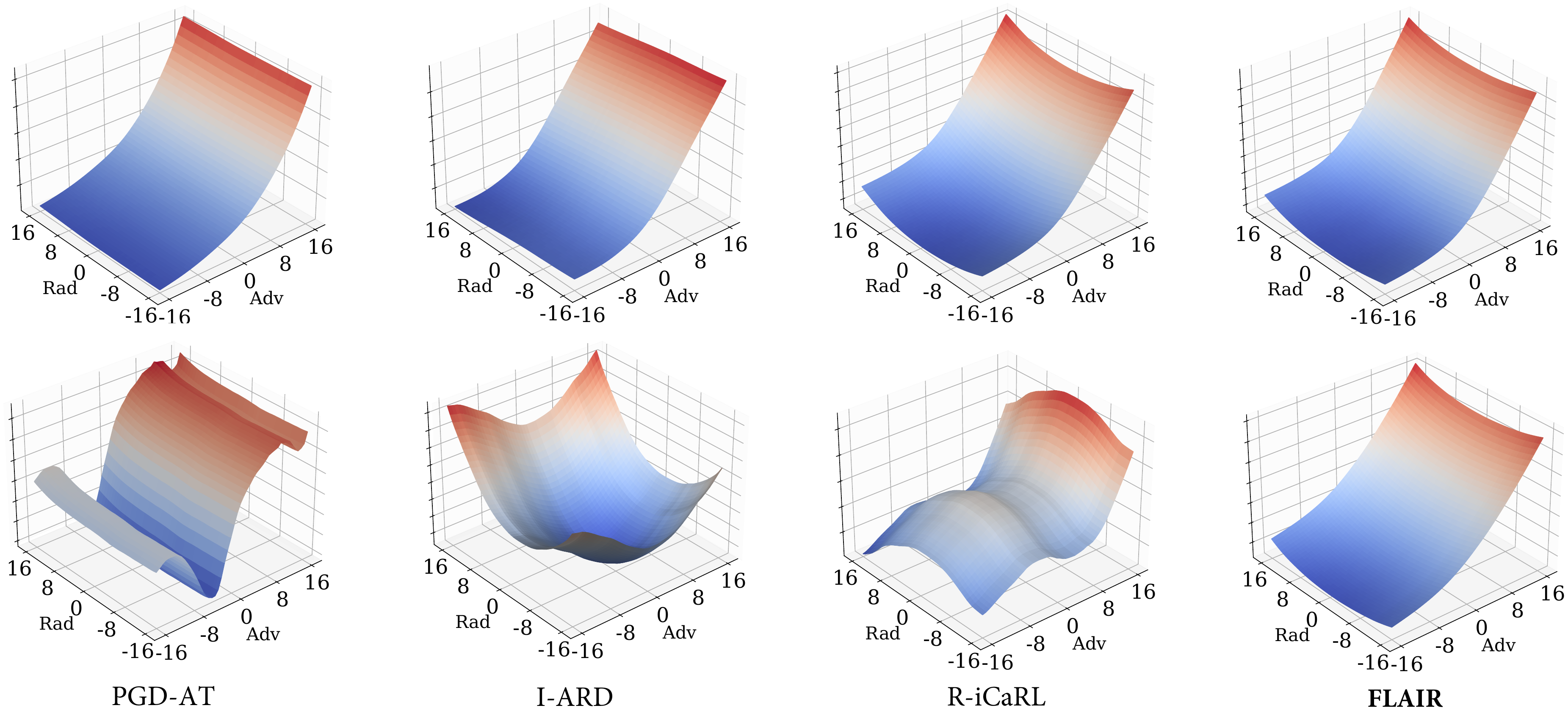}
  \caption{Loss landscape for each method at the beginning task $t$ (top row) and after learning task $t$ (bottom row).}
    \label{fig:visualization_loss_supp}
\end{figure*}
We further plot the loss landscape for flatness across other baselines using the visualization approach \cite{park2021reliably} in \Cref{fig:visualization_loss_supp}
We plot cross-entropy loss projected in two directions: adversarial and random direction.
Comparing our method with other methods, we find that our approach effectively preserves flatness.

\section{Training Details}
\subsection{Datasets} 
We introduce datasets that we utilized in experiments.

\subsubsection{CIFAR-10} \cite{krizhevsky2009learning} is a widely recognized benchmark dataset in machine learning and computer vision, comprising 60,000 32x32 color images categorized into 10 distinct classes: airplane, automobile, bird, cat, deer, dog, frog, horse, ship, and truck. The dataset is split into 50,000 images for training and 10,000 images for testing, providing a standard platform for evaluating image classification algorithms. Each class contains 6,000 images, and the dataset is commonly used for training and benchmarking various models. \textit{Split CIFAR-10 (S-CIFAR10)} is a modified version of CIFAR-10 where the dataset is divided into five tasks, each containing two classes based on the original class order. This partitioning is intended to facilitate incremental learning studies by progressively introducing new tasks and evaluating how models adapt to learning new class pairs while retaining knowledge of previously learned classes.

\subsubsection{CIFAR-100} \cite{krizhevsky2009learning} extends the CIFAR-10 dataset by offering a more challenging benchmark with 100 classes instead of 10. Each class is represented by 600 32x32 color images, resulting in a total of 60,000 images, split into 50,000 training images and 10,000 testing images. The classes in CIFAR-100 are organized into 20 superclasses, with each superclass containing five related classes. This hierarchical structure provides a more nuanced classification task compared to CIFAR-10. \textit{Split CIFAR-100 (S-CIFAR100)} further segments this dataset into ten incremental tasks, each encompassing ten classes.

\subsubsection{SVHN} \cite{SVHN} is a real-world dataset consisting of images of house numbers collected from Google Street View, containing a total of 10 classes, corresponding to the digits 0 through 9. It includes 73,257 images for training and 26,032 images for testing, with each image being a 32x32 pixel color image featuring one or more digits. Unlike the CIFAR datasets, SVHN presents a more complex and varied backdrop, with numbers appearing in diverse orientations, backgrounds, and sizes. \textit{Split SVHN (S-SVHN)} divides this dataset into five incremental tasks, each consisting of two classes.

\subsubsection{Tiny ImageNet} \cite{le2015tiny}, a subset of the larger ImageNet dataset, contains 200 classes with 500 training images and 50 validation images per class, all resized to 64x64 pixels. It offers a diverse range of visual contexts, maintaining complexity while being more manageable than the full ImageNet dataset. \textit{Split TinyImageNet (S-TinyImageNet)} divides this dataset into ten incremental tasks, each with 20 classes.

\subsection{Settings} We construct new baselines for ARCIL by applying adversarial training on those incremental learning.
In all cases of the methods, we employ ten steps of PGD with $\epsilon_{\infty} = 8/255$ and $\alpha = 2/255$ to perturb training images.
Furthermore, we train each task for 50 epochs for S-CIFAR10, and 100 epochs for both S-CIFAR100 and S-TinyImageNet.
The initial learning rates vary for each method, and we employ the same decreasing scheduler, reducing it by a factor of 10 at epochs 24, 31, and 40 for S-CIFAR10, and 48, 62, and 80 for both S-CIFAR100 and S-TinyImageNet for each incremental task.
We utilize SGD as the optimizer with no momentum and weight decay of $1 \times 10^{-5}$.
The batch size is 64,64,128, and 128 for S-CIFAR10, S-CIFAR100, S-SVHN, and S-TinyImageNet.
The initial learning rate is set to 0.1, 0.5, 1.0, and 1.0 S-CIFAR10, S-CIFAR100, S-SVHN and S-TinyImageNet.
Exceptionally, in R-EWC-on, a batch size of 32 was used because performance degraded with larger batch sizes.
Furthermore, for R-DER and R-DER++, the learning rate was set to 0.03.
The training details are consistent across different methods unless otherwise specified.

We conducted all experiments using an NVIDIA GeForce RTX 3090. The software environment was based on PyTorch, running on a Linux operating system. For most experiments, the memory usage was within 6GB, while for S-TinyImageNet, it was within 20 GB.

\subsection{Hyperparameters} For TRADES and MART, regularization parameter $\alpha$ is set to 6.
For I-ARD, I-RSLAD, and I-AdaAD, $\alpha$ and $\beta$ are set to 1.
For R-LwF, $\alpha$ is set to 1.
For R-DER, $\alpha$ is set to 0.3, and for R-DER++, $\alpha$ and $\beta$ is set to 0.1 and 0.5.
For R-LUCIR, $\alpha$ and $\beta$ is set to 0.3 and 0.5.
For FLAIR, we set $\alpha$ and $\beta$ to 0.5 and 2.

\section{Main Algorithm}
\begin{algorithm}[h]
\caption{Main Algorithm}
\label{alg:algorithm}
\textbf{Input}: Task Index $t$; current training dataset $S_{t}$; 
previous-step robust model $f_{t-1}$\\
\textbf{Output}: Updated robust model $f_{t}$
\begin{algorithmic}[1] 
\STATE Initialize $f_{t}$ with $f_{t-1}$
\IF{Augmentation = True}
\STATE $S_{t} \gets$  AuA($S_{t}$) \cite{cubuk2019autoaugment} \\
\hspace{2.5em} or RA($S_{t}$) \cite{cubuk2020randaugment}
\ENDIF
\FOR{epochs}
\FOR{each batch $(\boldsymbol{x}, y)$ in $S_{t}$}

\STATE $\boldsymbol{x}_{adv} = $ PGD$(\boldsymbol{x}, y ; f_{t})$ \cite{PGD}
\IF {$t=0$}
\STATE $l_{min} = l_{BCE} (f_t(\boldsymbol{x}_{adv}),\boldsymbol{1}_y)$
\ELSE
\STATE $l_{min} = \big[l_{BCE} ([f_t(\boldsymbol{x}_{adv})]^{t}_{t-1}, \boldsymbol{1}_y)$ \\
\hspace{3em} $+ \alpha \!\cdot\! l_{BCE}([f_{t}(\boldsymbol{x}_{adv})]^{t-1}_{0}, {f}_{t-1}(\boldsymbol{x}_{adv}))$\\
\hspace{3em} $+ \beta \!\cdot\! l_{FPD}(\boldsymbol{x}, \boldsymbol{x}_{adv} ; f_t, f_{t-1})\big]$

\ENDIF
\STATE $l_{min}$.backward()
\ENDFOR
\ENDFOR
\STATE \textbf{return} $f_{t}$
\end{algorithmic}
\end{algorithm}

In  \Cref{alg:algorithm}, we outline our approach. Here, $l_{FPD}$ and $L_{LAD}$ represent FPD and LAD, as described in the main paper.

\end{document}